\documentclass[journal]{IEEEtran}
\ifCLASSINFOpdf
\else
\fi
\usepackage{url}
\usepackage{times}
\usepackage{helvet}
\usepackage{courier}
\usepackage{amssymb}
\usepackage{amsmath}
\usepackage[caption=false,font=footnotesize]{subfig}
\usepackage[table]{xcolor} 

\usepackage{multirow}
\usepackage{hhline}
\usepackage{graphicx}
\usepackage[super]{nth}

\begin{document}
%
\title{Using Indirect Encoding of Multiple Brains to Produce Multimodal Behavior}
%
%
%

\author{Jacob Schrum, 
        Joel Lehman, 
        and~Sebastian Risi
\thanks{J. Schrum is with the Department
of Mathematics and Computer Science, Southwestern University, Georgetown,
TX, 78626 USA. e-mail: schrum2@southwestern.edu}
\thanks{J. Lehman and S. Risi are with the Center for 
Computer Games Research, IT University of Copenhagen, 
Copenhagen, Denmark. e-mail: \{jleh,sebr\}@itu.dk}}

\maketitle

\begin{abstract}
An important challenge in neuroevolution 
is to evolve complex neural networks with 
multiple modes of behavior. 
Indirect encodings 
can potentially answer this 
challenge. Yet in practice, indirect
encodings do not yield effective multimodal
controllers. Thus, this paper introduces
novel multimodal extensions to HyperNEAT, 
a popular indirect encoding.
A previous multimodal HyperNEAT approach
called situational policy geometry assumes
that multiple brains benefit from being embedded
within an explicit geometric space. However, experiments
here illustrate that this assumption
unnecessarily constrains evolution, resulting in
lower performance. 
Specifically, this paper introduces HyperNEAT extensions 
for evolving many brains without assuming geometric relationships
between them. 
The resulting Multi-Brain HyperNEAT can exploit 
human-specified task divisions to decide when
each brain controls the agent, or 
can automatically discover when brains should be used,
by means of preference neurons.
A further extension called module
mutation allows evolution to discover the number
of brains, enabling multimodal behavior
with even less expert knowledge.
Experiments in several multimodal domains 
highlight that multi-brain approaches are more
effective than HyperNEAT without
multimodal extensions, and 
show that brains without a geometric relation
to each other outperform
situational policy geometry. 
The conclusion
is that Multi-Brain HyperNEAT provides several 
promising techniques for evolving complex multimodal behavior.
\end{abstract}


%
\IEEEpeerreviewmaketitle

\section{Introduction}
\label{section:Introduction}
%
%
%
%
\IEEEPARstart{S}{uccess} 
in many AI domains
requires agents capable of complex
multimodal behavior, i.e.\ agents able to switch
between distinct policies based on
environmental context. 
Humans excel in this regard,
as they can switch fluidly between both physical
(sports, dancing, labor)
and intellectual 
(planning, writing, problem solving)
tasks. Such behavior is 
vital for a general AI agent, and
necessary for 
more focused agents as well, such as
robots, video game agents, and
agents in artificial life simulations.

One promising approach for creating
policies for agents is 
neuroevolution~\cite{local:stanley:ec02,floreano2008neuroevolution},
i.e.\ evolving artificial neural networks (ANNs). 
While there exist multimodal methods for
neuroevolution~\cite{local:calabretta:alife00,schrum:tciaig16}, 
many are \emph{direct encodings}, i.e.\ each component 
of an ANN is explicitly and distinctly encoded.
However, such direct encodings cannot exploit
regularities among inputs and outputs, and
do not scale
well to problems requiring large ANNs, motivating
\emph{indirect encodings} that can compactly
represent large networks. For these reasons,
indirect encodings capable of multimodal behavior 
are an important area of research. 

This paper thus proposes new extensions
to a popular indirect encoding called
HyperNEAT~\cite{stanley:alife2009}. 
A previous multimodal extension to
HyperNEAT is \emph{situational policy 
geometry}~\cite{dambrosio:iros2011},
an approach that creates
separate controllers for different 
situations defined by a human-specified
task division.
However, there are three problems with this
approach. First, situational policy geometry
requires there to be a geometric relationship
between different controllers (e.g.\
advancing and retreating are geometric opposites).
But in practice, an agent may need distinct policies 
that are not geometrically related.
Second, the human-specified task division
that is required imposes a burden of expert knowledge.
However, it is not always clear when different
modes of behavior should be used. Third,
the number of policies to generate must be set
in advance, requiring additional human 
knowledge.

A direct-encoded approach to learning multimodal behavior
without these limitations is
Modular Multiobjective NEAT 
(MM-NEAT~\cite{schrum:gecco14,schrum:tciaig16}).
MM-NEAT networks have several 
output modules, each of which defines 
a different policy. These policies
are not geometrically related,
and evolution
can decide when each policy should be
activated using preference neurons.
When combined with module 
mutation~\cite{schrum:tciaig12,schrum:tciaig16},
evolution can add modules as needed without
human intervention.

However, because MM-NEAT is a direct encoding, 
it cannot exploit regularities among inputs
and outputs. Further, because in direct encodings 
a network's parameter count is proportional to its
size (i.e.\ the curse of dimensionality), such
methods struggle to evolve complex ANNs; note that
this second advantage is not directly tested in
this paper but is a well-known property of 
HyperNEAT~\cite{hausknecht:tciaig14,gauci:aaai08,gauci:ppsn2010}.
The motivation for this paper's approach is thus to combine HyperNEAT 
with MM-NEAT to leverage the advantages of both: 
indirectly encoded controllers can better scale and
exploit domain regularities, 
while MM-NEAT allows evolution to create new modules
and discover how to arbitrate between them (without
assuming any geometric relationship between modules). 
The result is a system called
Multi-Brain HyperNEAT 
(MB-HyperNEAT)\footnote{Download at \url{southwestern.edu/~schrum2/re/mb-hyperneat.html}}.

MB-HyperNEAT is evaluated in four representative multimodal
domains, including two introduced in this paper.
The results indicate that policies unconstrained
by geometry outperform situational policy geometry.
Additionally, when preference neurons are used
to allow evolution to discover how and when to use
each brain, agents outperform standard HyperNEAT (without
multimodal extensions). In this way, MB-HyperNEAT
highlights the possible benefits from porting
previous multimodal approaches to indirect
encodings in a principled way.

\section{Background}
\label{section:Background}

This section discusses
previous approaches to evolving multimodal
behavior, and then describes HyperNEAT,
which the approaches in this paper extend
in a manner similar to yet distinct from situational
policy geometry, which is also explained.

\subsection{Evolving Multimodal Behavior}
\label{subsec:multimodal}

Because complex domains often
encompass many diverse and distinct 
subtasks, agents in such domains must
exhibit multimodal behavior
to succeed. For example, 
being able to defend and advance the
ball in soccer, or exhibiting
offensive and defensive behavior in a video game.
The Universal Approximation Theorem~\cite{haykin:book99}
indicates that a properly configured neural network
can theoretically exhibit any behavior,
which includes multimodal behavior.
However, in practice such behavior is more
effectively produced by modular
networks, as the examples below demonstrate.

Modular ANNs correspond to structures seen in
biology, and can represent distinct policies for the
subtasks within a multimodal domain. 
For these reasons, modular ANNs 
are an active area of 
research~\cite{kashtan:nasusa2005,clune:royal2013,huizinga:gecco14}.
Most multimodal approaches either implement evolutionary mechanisms that encourage 
modularity~\cite{kashtan:nasusa2005,clune:royal2013} or explicitly divide 
ANNs into modules that can specialize to different 
tasks~\cite{nolfi:adaptivebehavior96,local:calabretta:alife00,schrum:tciaig16,schrum:gecco2015}. 

Another approach that easily allows for
multiple modes of behavior is to use several
distinct networks to make decisions.
An example of this approach is Neural Learning Classifier 
Systems \cite{howard:cec2010,hurst:alife06,dam:tkde2008},
in which a single agent is controlled by a population of neural networks,
subsets of which activate to handle particular situations.
During learning, activated networks are generally modified according to
a rule similar to that used in temporal difference learning.
Individual networks also accrue fitness whenever activated,
and a genetic algorithm is periodically or probabilistically
used to allow offspring of fitter networks to replace less fit networks.

Multiple distinct networks can also be combined
to control a single agent if a human
trains or evolves each component separately,
and then combines them in a hierarchical configuration.
For example, Togelius's evolved subsumption 
architecture~\cite{togelius:jifs04}
was used in
EvoTanks~\cite{thompson:cig2009}
and Unreal Tournament~\cite{vanhoorn:cig2009}.
Lessin et al.\ used the principles
of encapsulation, syllabus, and pandemonium 
to evolve complex behavior for virtual 
creatures~\cite{lessin:gecco13, lessin:alife14}.
These approaches still require a programmer to
divide the domain into constituent tasks and
develop effective training scenarios 
for each task. 


The primary inspiration for this paper is 
individual networks divided into explicit modules.
This approach allows an agent to have distinct output 
modules corresponding to
separate policies meant to be used in different situations.
Networks can either have a human-designated
module for each task, as with
Multitask Learning~\cite{caruana:icml93},
or evolution can discover when and how
to use each module through special
\emph{preference neurons}. When an
ANN with preference neurons is
evaluated, the module whose preference neuron has 
the highest activation determines the final output. 
Preference neurons can also
be combined with module mutation~\cite{schrum:tciaig12}, 
an operation that
adds new modules, freeing the experimenter from
fixing the number in advance.

This paper adapts these ideas to
extend HyperNEAT, which is described next.

\subsection{HyperNEAT}
\label{subsection:HyperNEAT}

\begin{figure*}[t!]
  \centering 
\makebox[\textwidth]{
\subfloat[Standard HyperNEAT (One Brain)]{
  \label{fig:1M}
  \includegraphics[width=0.73\columnwidth]{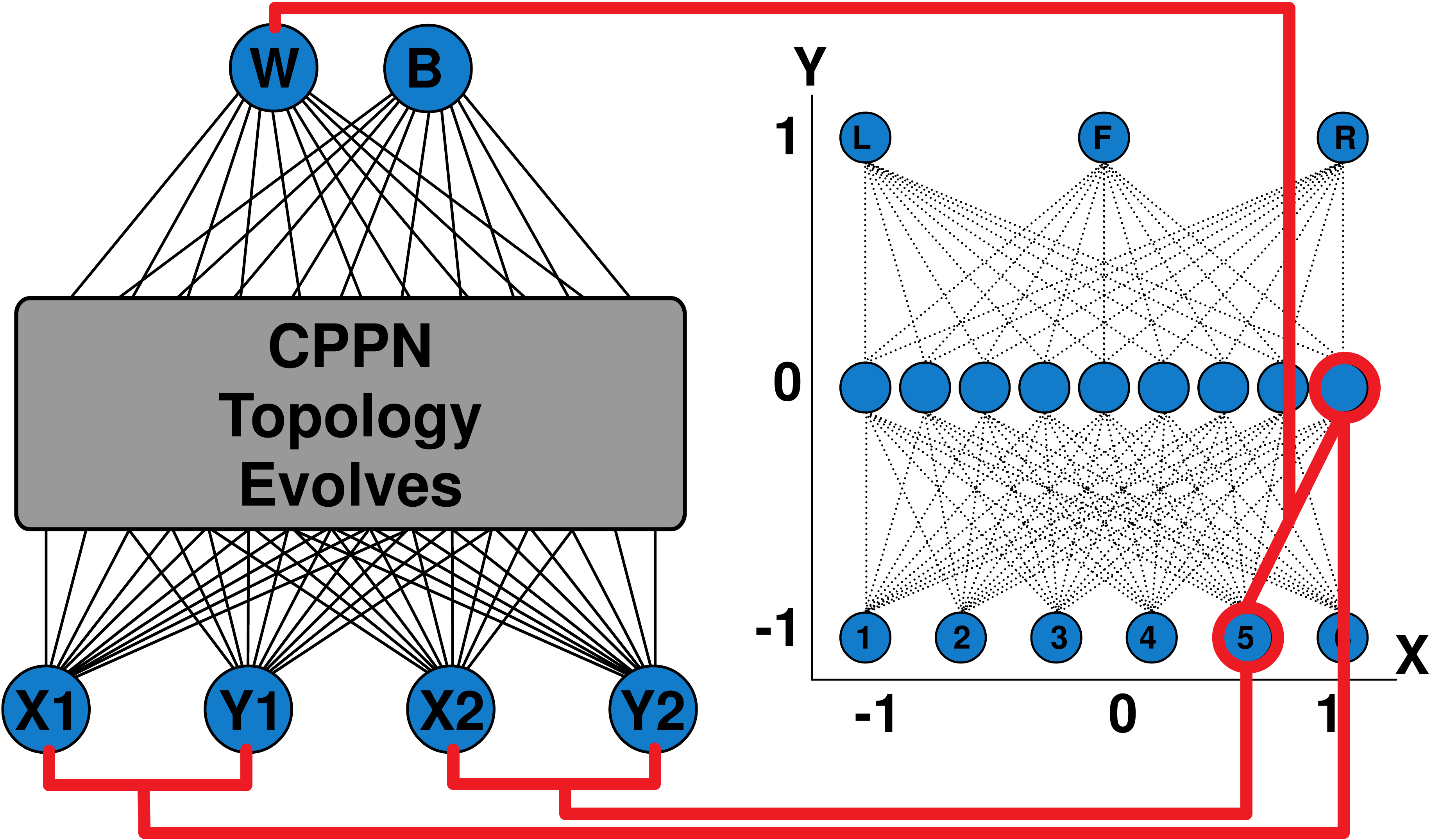}} 
\subfloat[Situational Policy Geometry]{
  \label{fig:SPG}
  \includegraphics[width=\columnwidth]{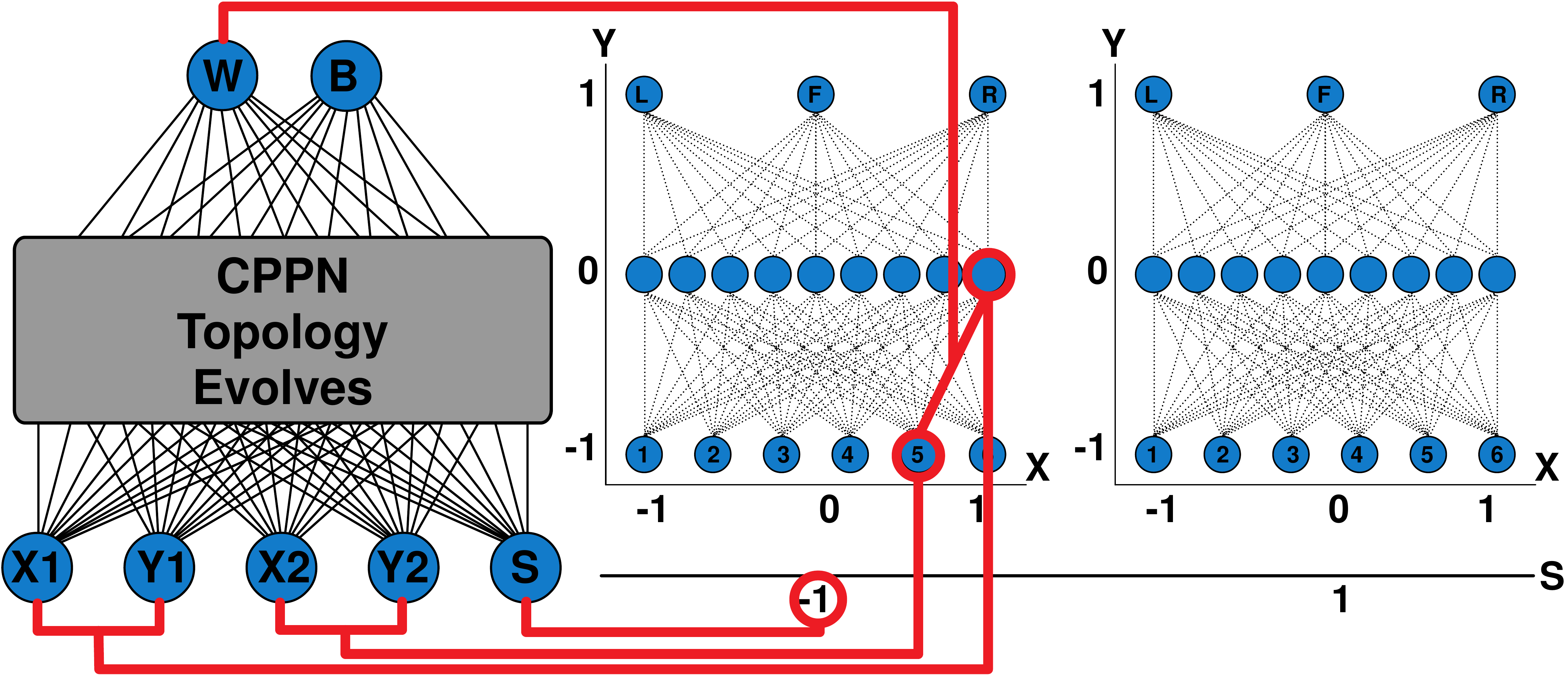}} 
}
\makebox[\textwidth]{
\subfloat[Multitask]{
  \label{fig:MT}
  \includegraphics[width=\columnwidth]{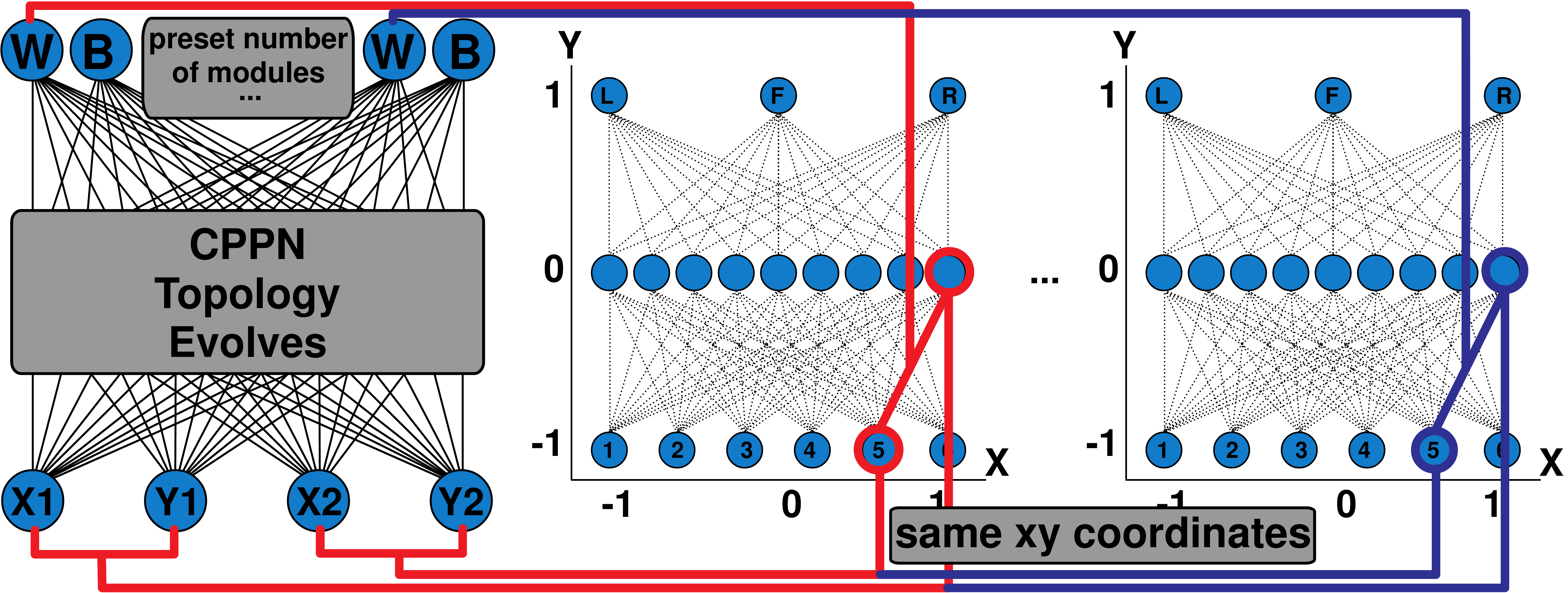}}
\subfloat[Preference Neurons]{
  \label{fig:Pref}
  \includegraphics[width=\columnwidth]{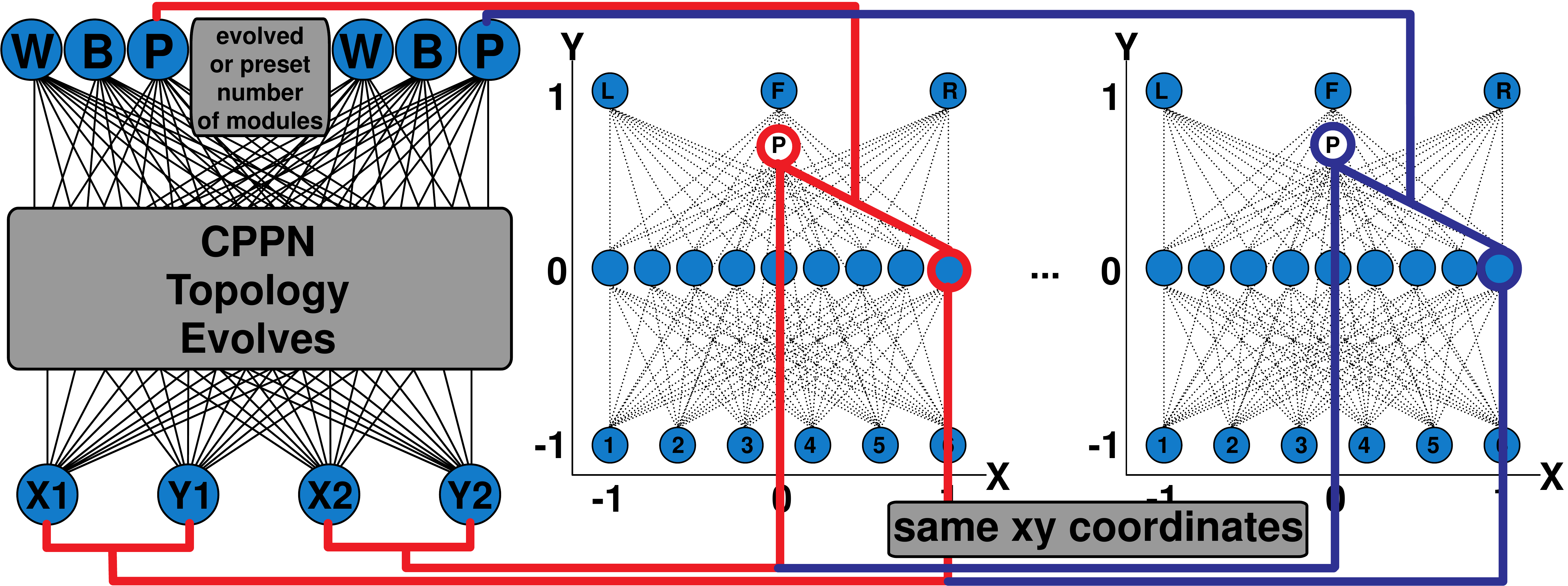}} 
} 
\caption{\small {\bf Methods for Generating Substrate Brains.} 
{\normalfont Each figure
shows a CPPN on the left and the corresponding substrate brain(s) it 
produces on the right. Each successive figure adds new features: 
situational policy geometry encodes multiple brains, 
multitask encodes
brains without geometric relationships, and 
preference neurons specify when
to switch brains.
\protect\subref{fig:1M} {\bf Standard HyperNEAT~\cite{stanley:alife2009}}: 
The CPPN creates a single-brain substrate.
Dotted lines in the substrate indicate which neurons can be connected.
For each possible connection, the xy-coordinates of both neurons are input
into the CPPN. The CPPN output $W$ determines 
whether a connection is created, and if it is, what its weight value will be.
The $B$ output determines a fixed bias from $(0,0)$ to each substrate neuron
(not shown).
\protect\subref{fig:SPG} 
{\bf Situational policy geometry~\cite{dambrosio:iros2011}}: 
CPPNs have an additional input $S$, which defines
a space of possible brains. The substrate contains 
a separate brain for each value of $S$ the CPPN is queried with.
Brains for $S$ values of $-1$ and $1$ are shown. 
The next two approaches are new to this work:
\protect\subref{fig:MT}
{\bf Multitask}: The CPPN has no additional inputs, 
but instead has a separate group
of outputs for each distinct brain.
Each brain is encoded independently, rather than
being embedded along an explicit $S$ dimension.
When the CPPN
is queried with a pair of neuron coordinates,
each output module supplies the corresponding connection 
weight for a different brain. 
\protect\subref{fig:Pref}
{\bf Preference neurons}:
When using preference neurons,
the number of modules
can either be preset or evolved using
module mutation (Figure~\ref{fig:MM}).
Preference neuron brains are generated 
similarly to multitask brains,
except that each output
module of the CPPN has an additional neuron $P$
that is queried only when the postsynaptic neuron
whose xy-coordinates are being input
is a preference neuron. 
The preference neuron (white) of each brain
is embedded in the substrate at
coordinate $(0,0.8)$, and is 
queried for potential connections
to all neurons in
the hidden and input layers.
All brains are activated on each time step, but only
the one with the highest preference neuron output determines
the agent's behavior on a given time step.}}
\label{fig:allMethods}
\end{figure*}

Hypercube-based Neuro-Evolution of 
Augmenting Topologies (HyperNEAT~\cite{stanley:alife2009})
is an extension of NEAT~\cite{local:stanley:ec02}, 
a direct encoding that
evolves arbitrary-topology ANNs
through mutations that gradually 
complexify networks.
HyperNEAT uses NEAT as a mechanism
to specify connectivity patterns
across an indirectly-encoded \emph{substrate}
ANN. 

Such connectivity patterns
are represented by Compositional Pattern Producing 
Networks (CPPNs), which differ from 
NEAT networks
in that (1) different neurons can have different
activation functions
chosen from a hand-designed set,
and (2) they are intended to be
queried repeatedly 
across a coordinate space
to produce a pattern.
The activation function set
includes functions 
that can produce useful ANN connectivity patterns,
e.g.\ symmetry and repetition.
Among other applications, 
CPPNs can be queried across 2D space to
produce images~\cite{secretan:ecj2011}, or across 4D
space to produce ANNs as in
HyperNEAT.

For a CPPN to generate
a substrate network, the 
substrate must be embedded within a geometric 
space (Figure~\ref{fig:1M}).
To generate the connectivity of the
substrate, each set of possible
connections is queried through the CPPN.
In particular, for each connection the 
 coordinates of the source and target neurons
in the substrate are provided as input to the CPPN. 
The primary CPPN output is then interpreted as
the connection weight between these
two neurons, although the connection is created only if
this value is greater than a threshold value.
There is also a separate CPPN output
queried once per target neuron 
(the other input coordinates are $(0,0)$) 
that specifies
the fixed bias of that neuron.

The geometric layout of the substrate
is designed by the experimenter. This
design specifies
how many neurons are in each layer, and
whether neurons are input, output, or
hidden neurons.
Ways of automatically configuring
this substrate exist~\cite{risi:alife12},
but are not necessary for the experiments of this paper.
The geometric embedding 
of the ANN in the substrate is
both a strength and a weakness of HyperNEAT.

On one hand, a geometric embedding allows a CPPN
to exploit task-relevant geometry.
For example, because there is often a meaningful 
relationship between the geometry of an agent 
(the placement and orientation of its
sensors and effectors)
and its policy (e.g.\ sensory input from a 
particular direction might encourage moving
away from or toward that direction), it is common to align
the geometry of the substrate with that of the 
agent. In this way, HyperNEAT can exploit geometry to create
policies with similar regularities. 
In contrast, a direct encoding might need to
learn the underlying holistic pattern separately for each
sensor and effector.

On the other hand,
sensors or actuators that have no obvious
geometric interpretation cannot easily be
embedded in a principled way. Researchers have
addressed this challenge by adding
new dimensions or substrates to handle
different sensor modalities~\cite{drchal:icann2009,pugh:gecco2013,hausknecht:tciaig14}.

This idea can be extended to create completely
independent networks, and has been previously
used by the situational policy geometry approach,
described next.

\subsection{Situational Policy Geometry}
\label{subsection:SPG}

Because CPPNs can be fed
neuron coordinates from a continuous
space, they can
generate arbitrarily large and complicated 
substrate ANNs, which in theory can yield
arbitrarily complex behavior, including
multimodal behavior.
However, in practice an easier way to realize 
multimodal behavior is with several smaller networks, 
rather than with a single large one. 

An existing implementation of this approach
in HyperNEAT is called
situational policy 
geometry~\cite{dambrosio:iros2011}.
Agents have distinct brains for
different situations, but it is
assumed that the
brains share an underlying geometric relationship.
That is, CPPNs have an additional situation input,  
allowing multiple policies to be generated to
deal with different situations.
For example, a situation input of -1 might
generate a policy causing a robot to advance 
in a maze, while an input of 1 might
create a policy for returning home.
In this approach, the decision of which policy to
use when must be specified in advance by the
experimenter (Figure~\ref{fig:SPG}).

Though sometimes effective, assuming policies will
have a geometric relation is overly limiting.
Therefore, the next section describes several new ways
to create multiple brains through HyperNEAT, without
assuming such a geometric relationship.


\section{New Approaches Using Multiple Brains}
\label{subsection:ModularBrains}

This section presents three main
extensions to HyperNEAT, collectively
called MB-HyperNEAT.
Each idea is inspired by the direct-encoded 
MM-NEAT~\cite{schrum:tciaig16} approach. 
To make the following discussion as clear
as possible, the following terms are
defined: 

\begin{itemize}
\item A module, or output module, is a group of 
      related output neurons possessed by a CPPN.
	  It is responsible for creating a single brain within
      the substrate.
\item A brain is one artificial neural network created by
	  a CPPN. It exists within a substrate, and an agent
      may possess multiple brains.
\item An agent is an entity that takes action in an environment.
	  It may have multiple brains, but on any given time step,
      its action will be derived from only one of the brains.
      The agents in this paper are simulated robots.
\end{itemize}

These terms are used to describe 
three extensions to HyperNEAT:
multitask CPPNs, substrate brains with
preference neurons, and CPPNs subject to
module mutation.

\subsection{Multitask CPPNs}

Multitask networks were first proposed
by Caruana~\cite{caruana:icml93} 
in the context of
supervised learning using neural networks and backpropagation. 
One network has multiple modules, where each module
corresponds to a different, yet related, task. 
Each module is trained on the data for the task to which it corresponds,
but because hidden-layer neurons are shared by all outputs,
knowledge common to all tasks can be stored in the weights of
the hidden layer. This approach speeds up supervised learning
of multiple tasks (or even just a single task of interest) 
because knowledge shared across tasks is only learned once
and shared, rather than learned independently multiple times.

Multitask networks can also be evolved to control
agents~\cite{schrum:tciaig12,schrum:tciaig16}, by
using separate output modules to solve different tasks,
or by manually assigning each module to a different
part of the task.

This paper uses the structure from multitask learning
to create multitask CPPNs, which 
have separate output modules for each brain they 
define (Figure~\ref{fig:MT}).
For every neural connection the CPPN queries,
an output from each module defines the weight of
that connection in a different brain. Thus,
multiple brains are defined, but the use
of separate CPPN outputs means that there 
need be no geometric relation between 
the policies encoded by each brain.

However, a human must still specify when each
brain is used. This limitation is overcome with
the addition of preference neurons, described next.

\subsection{Preference Neurons}

Preference neurons 
make module arbitration without 
human-specified task divisions possible.
In directly-encoded network modules with
preference neurons~\cite{schrum:tciaig12,schrum:tciaig16},
each module's preference neuron
outputs the network's relative
preference for using that
module. Whenever inputs are presented
to the network, the
module whose preference neuron output is the
highest is used to define the output
of the network.

In this paper, individual substrate
brains have preference neurons.
Each brain must be activated with the same
inputs, corresponding to the agent's sensors, 
on each time step, but only the brain with
the highest preference neuron output will
define the agent's behavior on each time step.
Because preference neuron behavior is
ultimately determined by an agent's genotype,
it is up to evolution to discover when to use each brain.

Preference neurons exist within each brain
substrate. However, it would be limiting
if the behavior of the preference neuron
were tied directly to the geometry of the
policy its brain exhibited. Therefore, CPPNs
for preference neuron brains have an additional
output for each module that is used
to define link weights entering the
preference neuron (Figure~\ref{fig:Pref}).
Links between all other neurons are defined
using the module's standard weight output, as in
multitask CPPNs. Use of separate CPPN outputs
for these two categories of neuron provides
evolution with the flexability to discover
agent policies that exploit certain
patterns and regularities of the domain,
while the behavior of preference neurons
may focus on different patterns within the domain.

Although this approach allows evolution
to determine which brain to use
on each time step, it is still the CPPN
that determines how many brains an agent
will have. In particular, the number of 
CPPN modules determines the number of substrate
brains, but in standard HyperNEAT, there is no
way to add additional output neurons.
However, groups of output module neurons
can be added by Module Mutation, described next.

\begin{figure}[t!]
  \centering 
\makebox[\columnwidth]{
\subfloat[Before Module Mutation]{
  \label{fig:MMbefore}
  \includegraphics[width=0.2\textwidth]{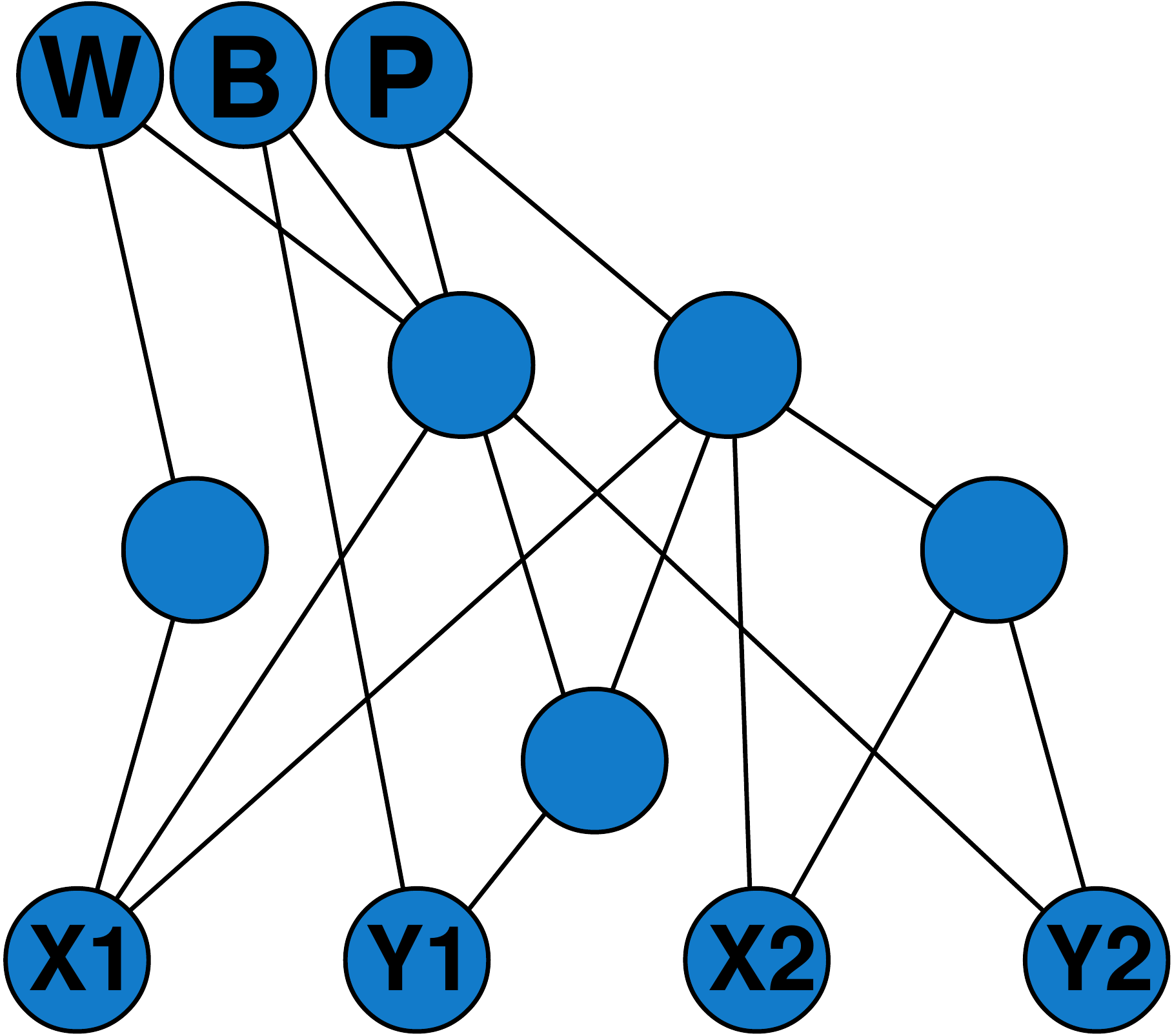}}
\subfloat[MM(P)]{
  \label{fig:MMP}
  \includegraphics[width=0.2\textwidth]{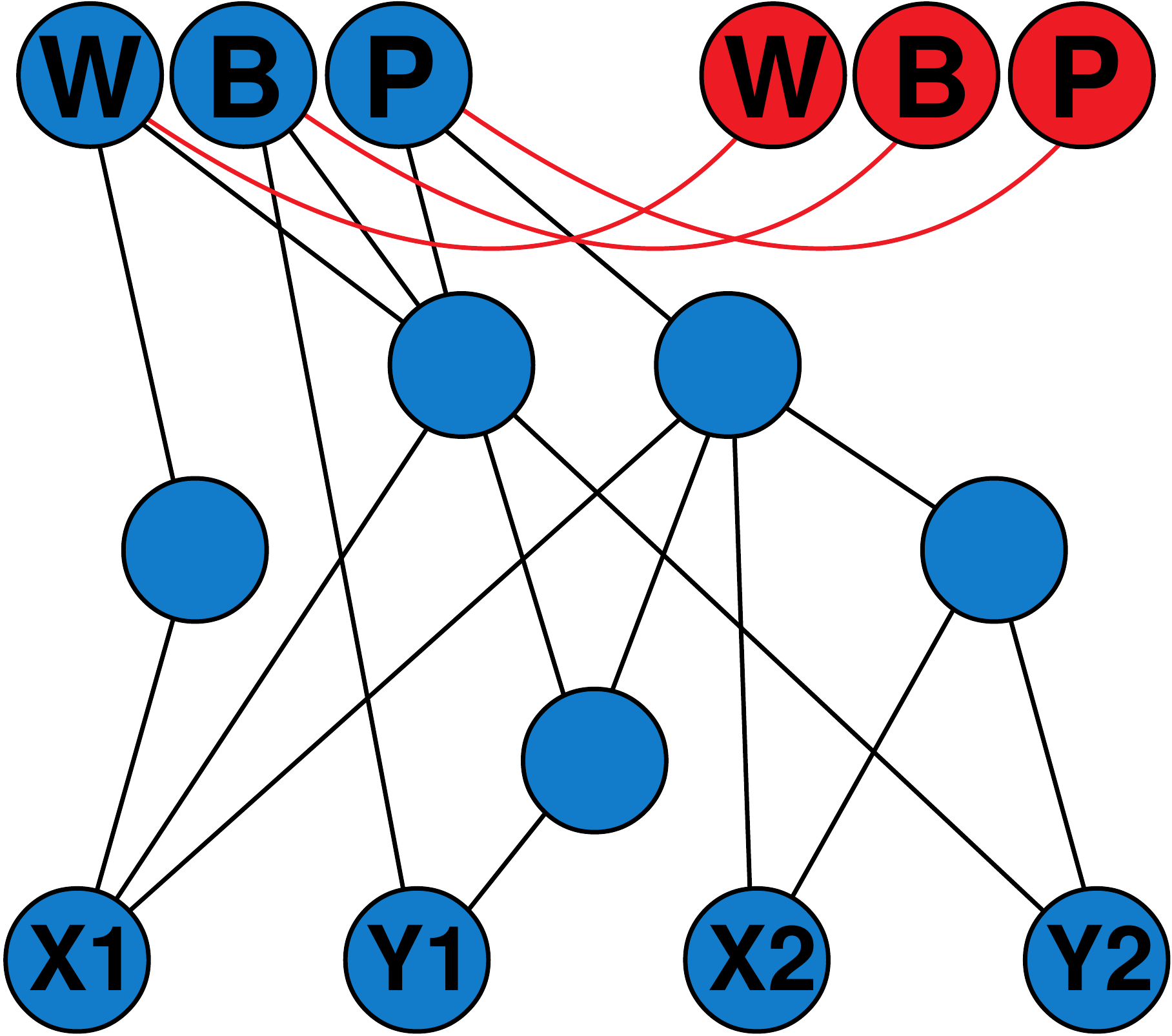}}
} \\
\makebox[\columnwidth]{
\subfloat[MM(R)]{
  \label{fig:MMR}
  \includegraphics[width=0.2\textwidth]{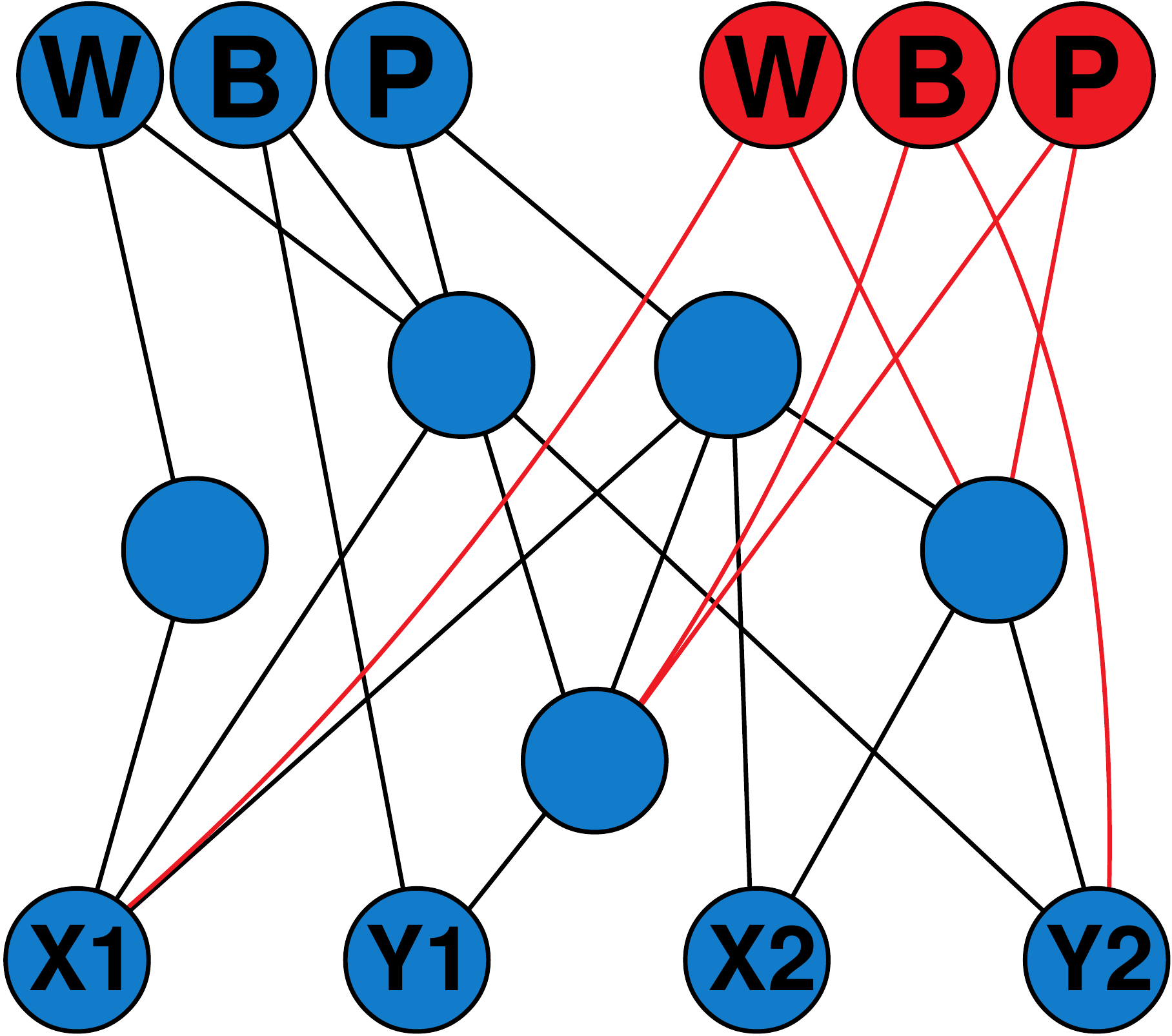}}
\subfloat[MM(D)]{
  \label{fig:MMD}
  \includegraphics[width=0.2\textwidth]{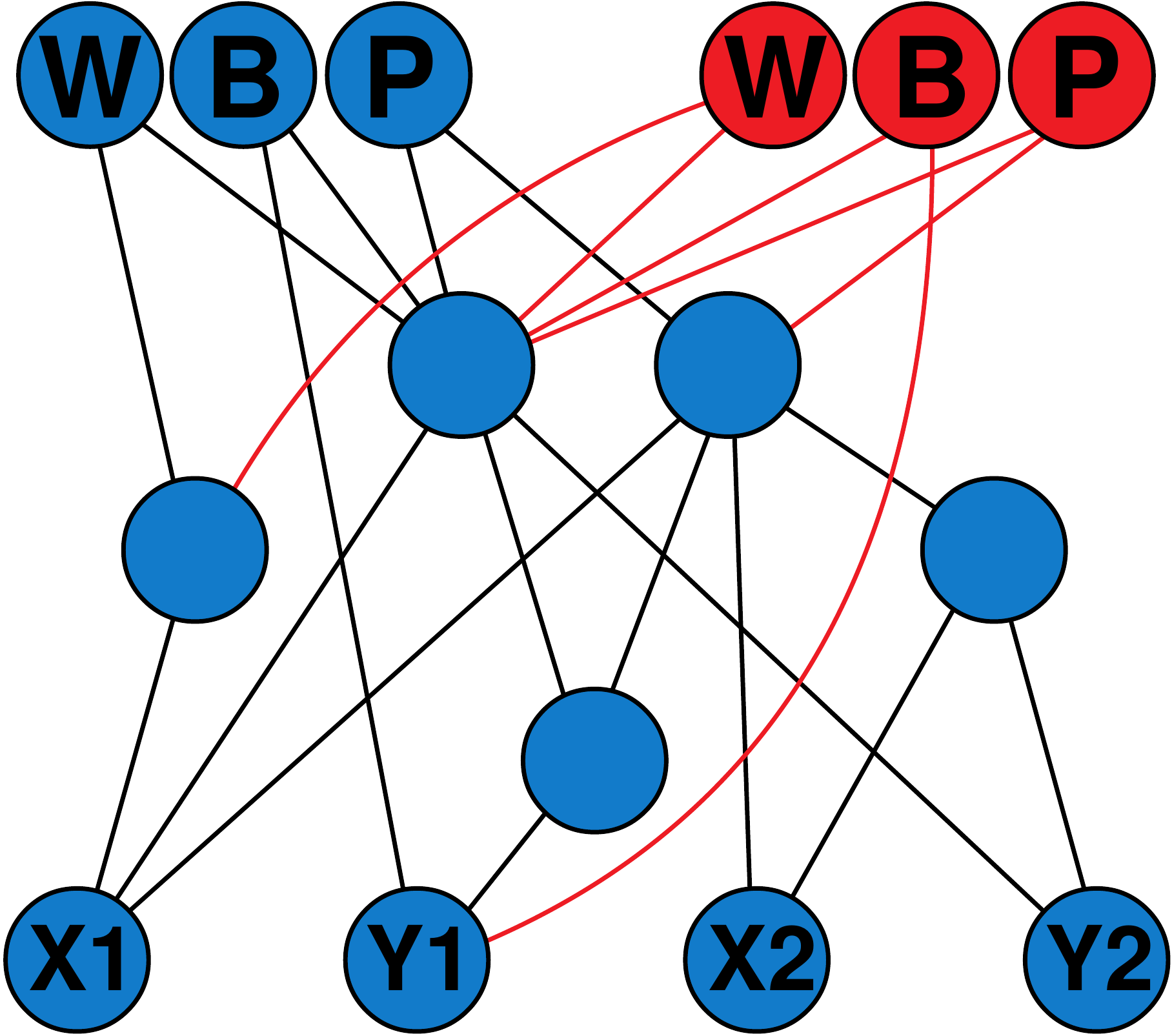}}
}
\caption{\small {\bf Three Types of Module Mutation.} 
{\normalfont Each form of module mutation
provides a different way for evolution to discover
how many modules to apply to
a problem.
\protect\subref{fig:MMbefore} At the start of
evolution each CPPN has a single module.
\protect\subref{fig:MMP} MM(P) creates a new
module with lateral connections from a randomly chosen previous 
module. The synaptic weight on each lateral connection is $1.0$.
The new module's behavior will depend on
the previous module, though activation functions
in the new neurons will cause the new module to have
slightly different outputs.
\protect\subref{fig:MMR} MM(R) creates connections from 
random source neurons with random synaptic weights leading into the
neurons of the new module. The number of 
incoming connections for each new neuron equals the
number of incoming connections to the corresponding neuron
of a randomly chosen previous module.
\protect\subref{fig:MMD} MM(D) creates a new
module that behaves identically to a 
randomly chosen previous module by creating
copies (source neuron and weight) of all incoming connections
to that module. Evolution can then cause the behaviors of
the duplicated module to diverge through mutations
in later generations.}}
\label{fig:MM}
\end{figure}

\subsection{Module Mutation}

Module Mutation~\cite{schrum:tciaig12,schrum:tciaig16} 
is any structural mutation operator
that adds a new output module to a neural network.
An indefinite number of modules may be added
in this way. Each network in an initial
population starts with a single module,
but as evolution progresses, different CPPNs
can possess different numbers of modules.

Each application of Module Mutation
to a CPPN adds an additional 
substrate brain to an agent.
Each substrate brain possesses
a preference neuron. As a result,
evolution is discovering which
brains to use while also discovering
the number of brains each agent should possess.

Several forms of module mutation are used in this paper: 
MM(P) for Previous, whose new module inputs
come directly from a previous output module,
MM(R) for Random, whose new module inputs
come from random sources in the network,
and MM(D) for Duplicate, whose new module
inputs are chosen to be the same as those entering
another module, thus duplicating the behavior
of that module. These approaches are more
thoroughly described in Figure~\ref{fig:MM}.

These three new approaches for creating
multiple brains for HyperNEAT agents
are compared against situational
policy geometry and single-brain agents
in several domains, which are described next.

\section{Experimental Domains}
\label{section:ExperimentalDomains}

\begin{figure*}[t!]
\centering 
\begin{minipage}[c]{0.66\textwidth}
\centering 
\subfloat[Team Patrol]{
  \label{fig:teamPatrolDomain}
  \includegraphics[width=3.5cm]{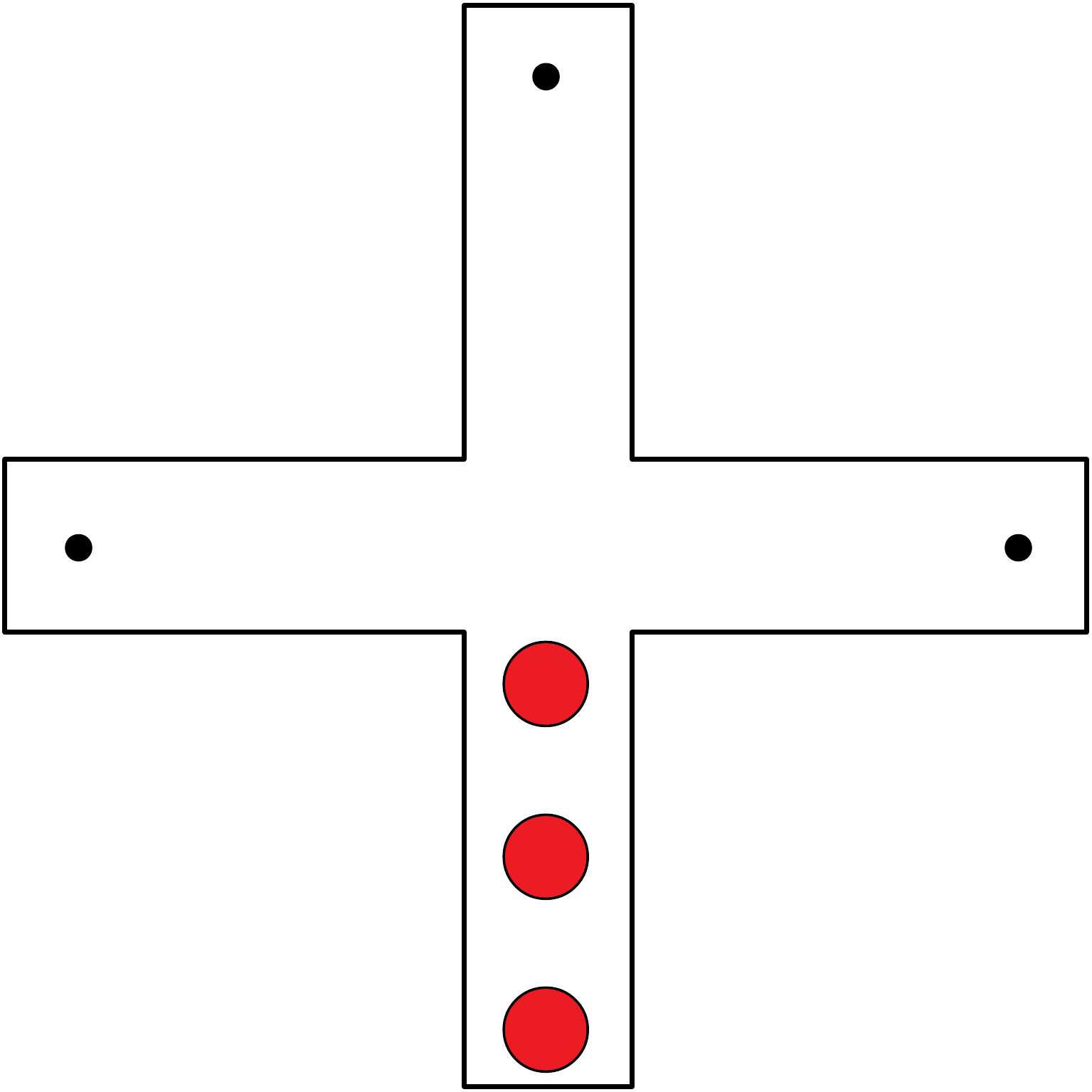}}
  \hspace{2cm}
\subfloat[Lone Patrol]{
  \label{fig:lonePatrolDomain}
  \includegraphics[width=3.5cm]{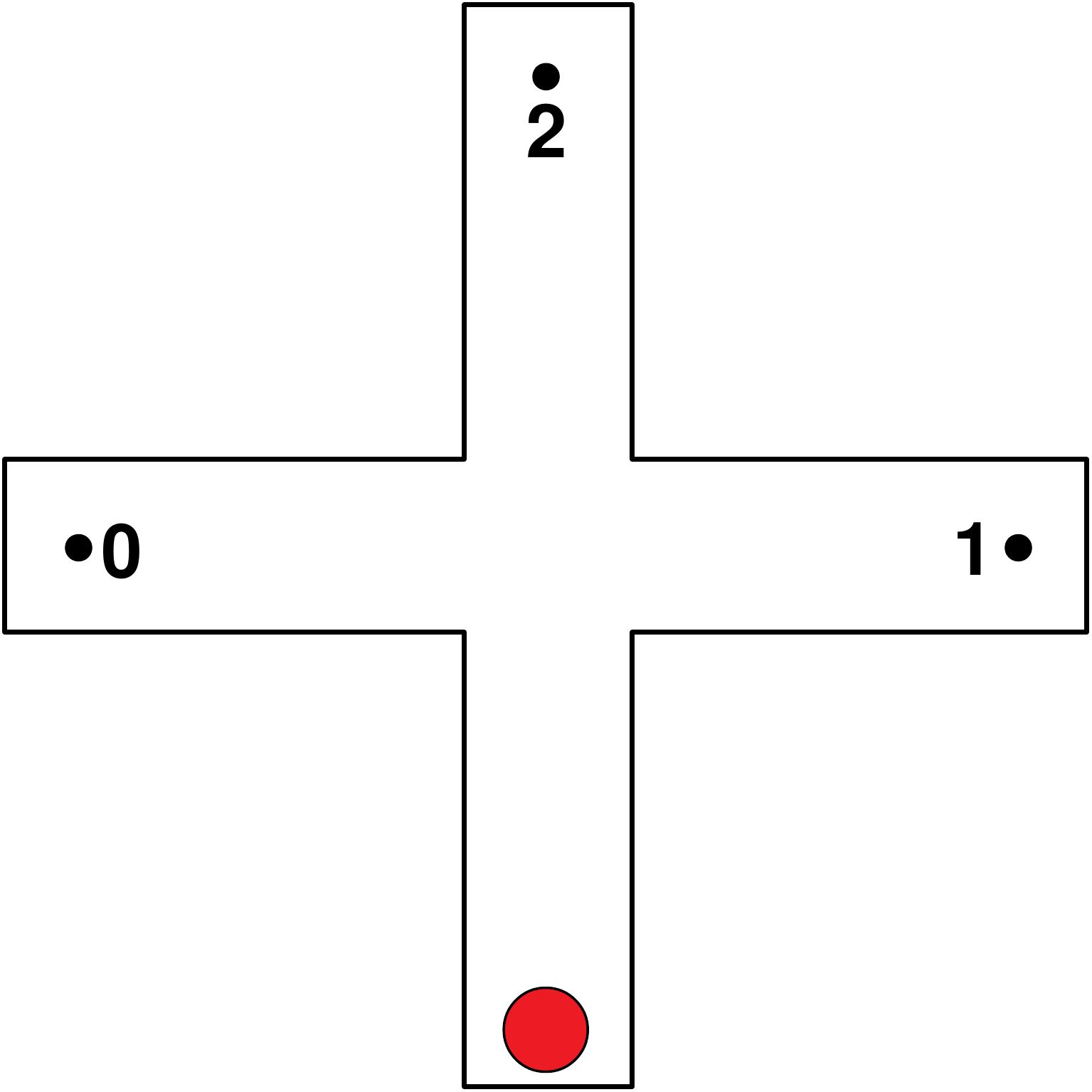}} 
  \\
\subfloat[Dual Task: Navigation]{
  \label{fig:dualTaskHallway}
  \makebox[3cm][c]{\includegraphics[width=2.45cm]{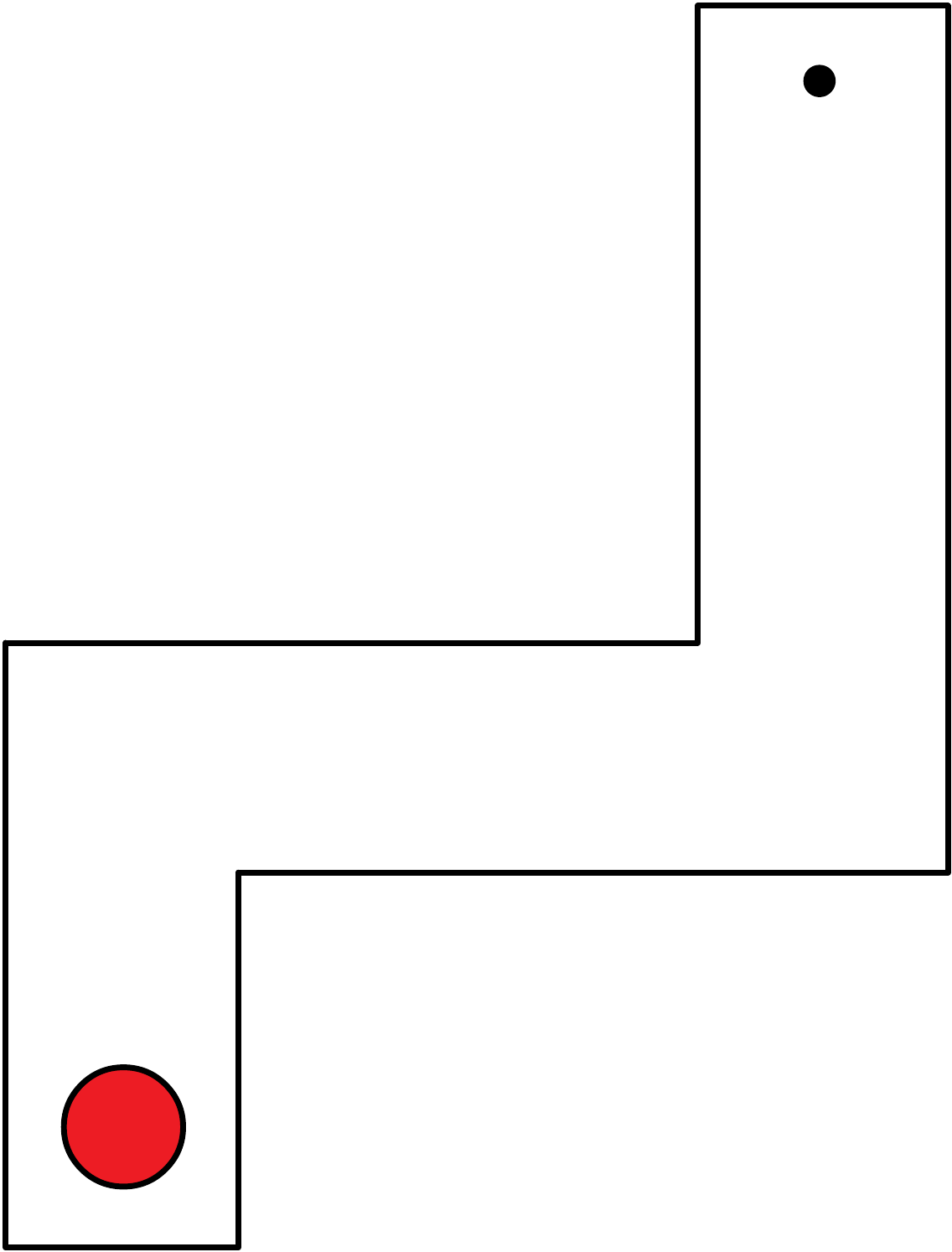}}} 
  \hspace{2cm}
\subfloat[Dual Task: Foraging]{
  \label{fig:dualTaskForage}
  \includegraphics[width=4cm]{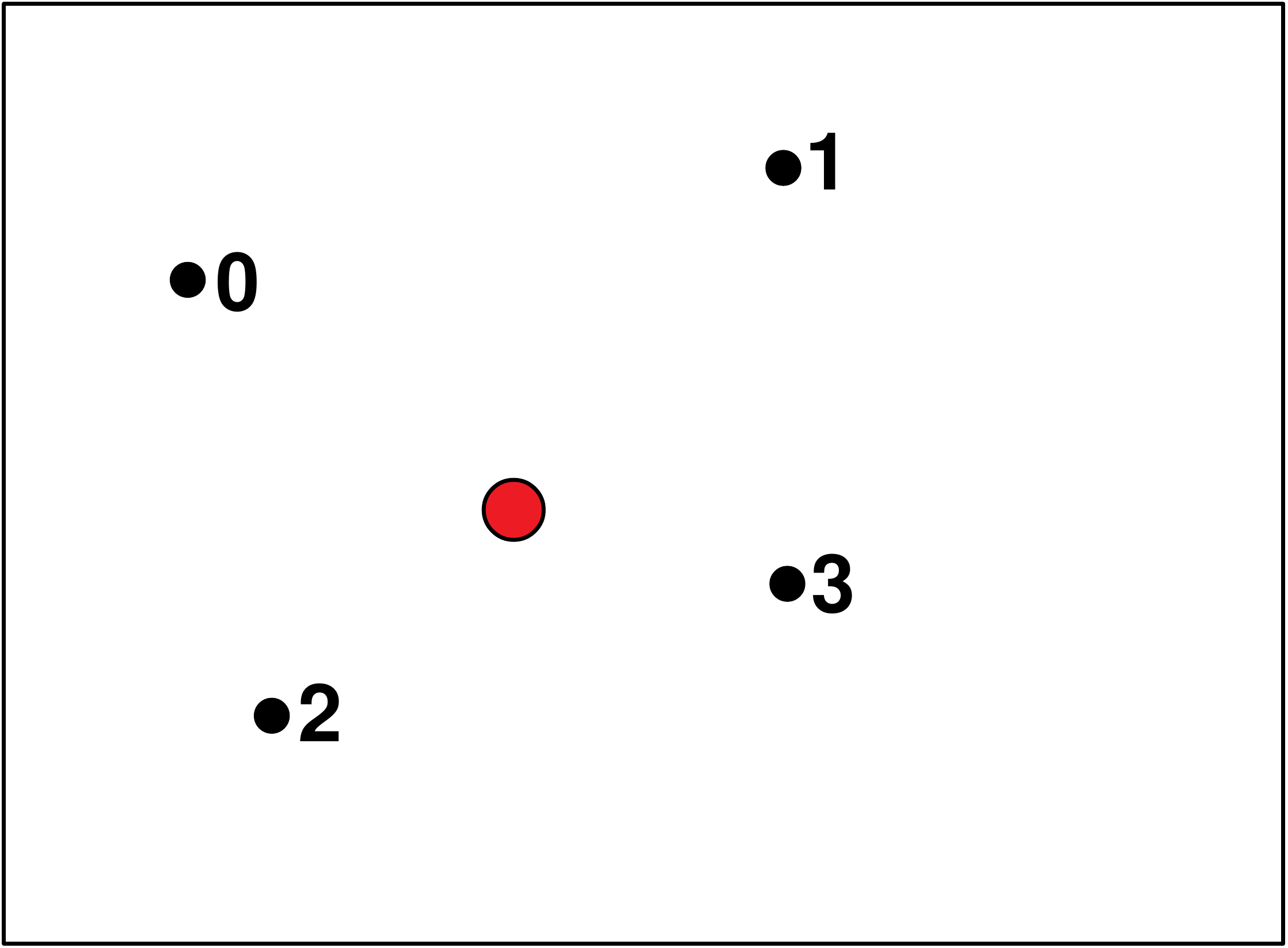}}
\end{minipage}%
\begin{minipage}[c]{0.33\textwidth}
\centering 
\subfloat[Two Rooms]{
  \label{fig:twoRoomDomain}
  \includegraphics[width=4.25cm]{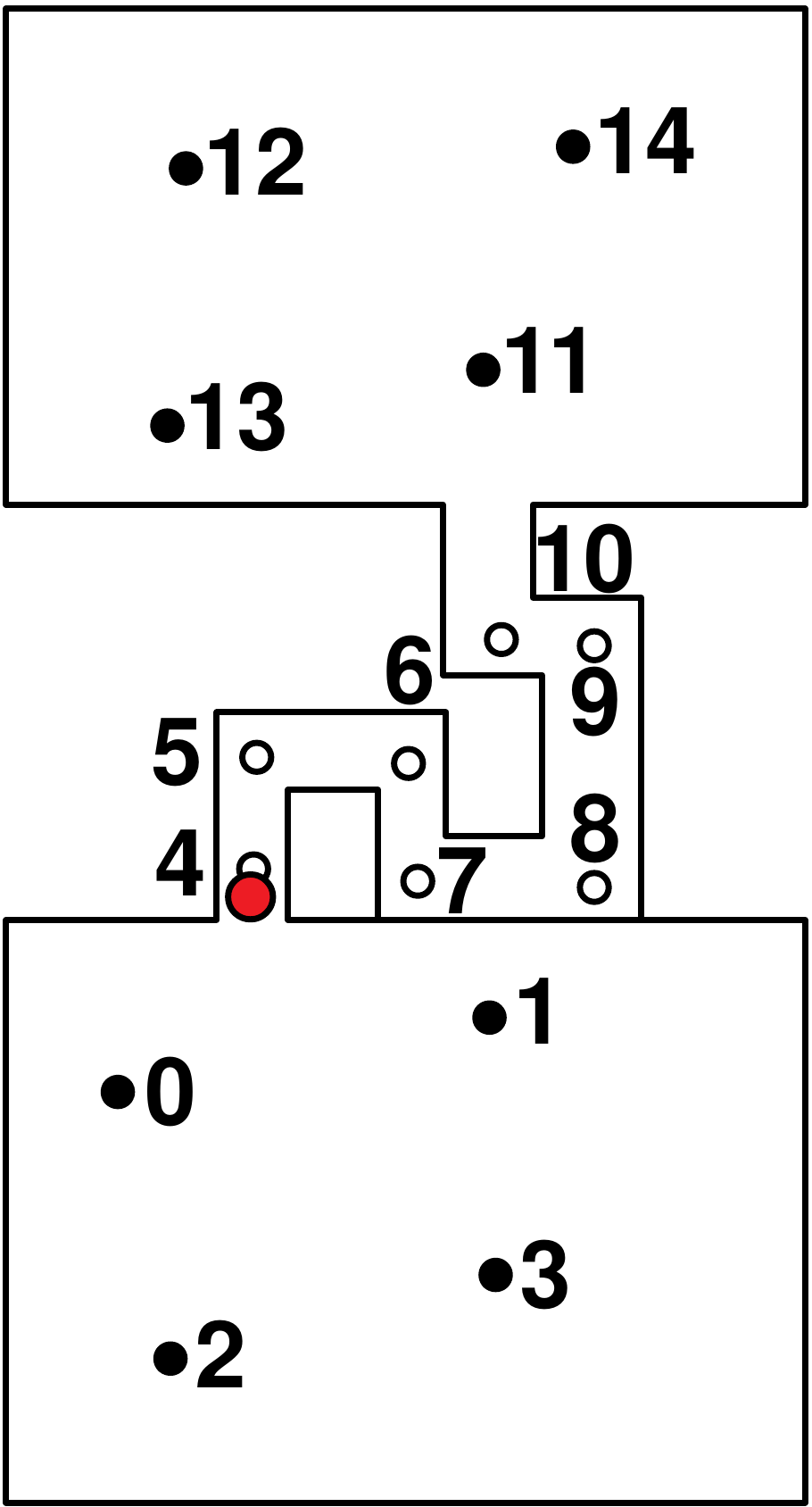}} 
\end{minipage}%
\caption{\small {\bf Domains Requiring Multimodal Behavior.}
{\normalfont Agents are evolved in four domains.
\protect\subref{fig:teamPatrolDomain} Team patrol requires
the three robots (red circles) to each visit 
a different waypoint (black dot) before returning to the
start point.
\protect\subref{fig:lonePatrolDomain} Lone patrol requires a
single robot to visit each waypoint in 
order before returning.
\protect\subref{fig:dualTaskHallway} The first 
dual task environment is a hallway, which evaluates
the robot's ability to navigate from one end to the other.
The same robot is then evaluated in the
\protect\subref{fig:dualTaskForage} foraging environment,
in which it must visit each waypoint in order.
\protect\subref{fig:twoRoomDomain} The two rooms
domain combines hallway navigation and foraging.
The robot must visit each waypoint in order,
clearing the lower room
first, and then traversing the hallway before moving
on to the upper room. The hollow pellets in the hallway
are invisible breadcrumbs that provide incremental
fitness benefit for progressing through the hallway, 
but cannot be perceived by the robot
as the waypoints can.
In aggregate, these domains feature different types
of task divisions, and thus make an interesting test suite
of problems to evaluate multimodal methods.}}
\label{fig:domains}
\end{figure*}

This section reviews two previous multimodal domains
and introduces two new ones (Figure~\ref{fig:domains}), 
which provide
a suite of representative tasks
to test the new approaches of this paper. 
These domains all
use simulated Khepera robots with
rangefinder sensors and three actuators:
one each for turning left, turning right,
and moving forward. 
On each time step, the robot
will perform whichever of the three
actions is most highly activated by
the controlling network brain.
In some domains,
robots also have pie-slice sensors
for detecting waypoints. 
The situation inputs used by situational
policy geometry in each domain are also specified.
These inputs always depend on a human-specified
task division that is also used by the multitask
approach.
Next, each
of these domains are motivated and
explained in turn.

\subsection{Team Patrol}
\label{subsection:teamPatrol}

The team patrol domain was originally used
to demonstrate the effectiveness of
situational policy geometry in 
HyperNEAT~\cite{dambrosio:iros2011},
and is thus an ideal comparison domain.

The domain is divided into two
tasks: advance, in which the three
robots spread out to the three 
segments of a room shaped like a plus sign, and
return, in which the robots must
return to their original starting 
positions (Figure~\ref{fig:teamPatrolDomain}).
Evaluation lasts 45 seconds, and each second
contains 30 time steps.
The task switch occurs at the midpoint
of the evaluation, regardless of whether
the robots reach their individual goals.

Fitness depends both on advancing to
the waypoints in each dead end, and
on returning home afterward.
While advancing, each robot is assigned
the closest waypoint as its goal, 
and while returning
the starting point is each robot's goal.
Every second each robot receives
a fitness increment equal to the
normalized distance from the robot to its goal.
In all domains of this paper, the
normalized distance is defined as
$\frac{D-d}{D}$, where $D$ 
is the maximum distance for the particular domain,
and $d$ is the current distance from the
robot to a point of interest. 

However, if the robot is
within $10$ distance units of its goal, it
is considered to have reached it and receives
a fitness increment of $1$. To further encourage success,
fitness is divided by $10$ if robots
do not move after the return signal is given,
or if not all waypoints are
successfully reached. Furthermore, the fitness
for proximity to the starting point is
divided by $100$ if the team of robots did
not actually reach all way points during
the advance stage of evaluation.
These specifics are rather complicated, but
are taken directly from the original publication
that introduced this domain~\cite{dambrosio:iros2011}.


Robots use six rangefinder sensors
tied to substrate inputs.
The sensors detect walls but not other
robots. In fact, the robots do not
physically interact because they are
meant to be deployed individually, despite
being evaluated simultaneously.
Each agent's substrate also has
nine hidden neurons, and
three outputs corresponding to the
left, forward, and right actions.
Because each agent on the team must
behave differently, each has its own brain(s).
This is accomplished using multi-agent HyperNEAT~\cite{dambrosio:aamas2010},
in which an additional input to the CPPN defines a team
dimension along which brains can vary. Each agent is
assigned a separate coordinate in this 
dimension ($-1$, $0$, or $1$).

For situational policy geometry,
two separate brains are generated for
each agent, through situation inputs of -1 and~1, 
as described in the example from section~\ref{subsection:SPG}.
One brain controls the agent during the advancing stage,
while the other is used during the retreating stage. 
When either there is only one 
brain per agent, or when using  preference neurons,
a separate situation input is required in
the substrates for each brain (at coordinates
$(0,-0.8)$). This input tells the brains 
whether they should currently
be advancing or retreating.

\subsection{Lone Patrol}
\label{subsection:lonePatrol}

Because agents must cooperate to solve 
the team patrol domain, it conflates the challenges
of multiagent coordination and multimodal
behavior. Thus, the lone patrol domain is
introduced to isolate
the multimodal aspect of team patrol.

This goal is accomplished by placing 
only a single robot 
in the same environment.
This robot is responsible for visiting
all branches of the plus sign 
(Figure~\ref{fig:lonePatrolDomain}). 
To add to the domain's challenge, 
the robot must visit the branches
in an order requiring the central 
four-way intersection to be
handled in three different ways: turning left,
going straight, and turning right. 
For situational policy geometry, 
the situation inputs -1,~0,
and~1 correspond to brains for 
these three behaviors,
which switch whenever the agent
reaches a waypoint.

The fitness function 
encourages the robot to reach each of
the waypoints as fast as possible in sequence.
On every time step fitness is incremented 
by the normalized distance 
from the robot
to its next goal.
However, there is an additional fitness increment 
of~$1$ per waypoint that has
already been reached. Therefore, the robot receives
increased fitness per time step for reaching  
additional waypoints. If the robot
has visited all waypoints 
and returned home, the fitness increment 
is $4$ ($1$ per waypoint) on each remaining
time step. Because it takes
an individual robot longer to visit all ends of
the plus sign, the evaluation time is 80 seconds.
though there are still 30 time steps per second.


Both of the domains described so far have
subtasks with clear geometrical relationships.
Therefore, one might expect situational
policy geometry to perform well in these domains. 
However, many
domains have no clear geometrical subdivision.
The next domain provides such an example.

\subsection{Dual Task}
\label{subsection:dualTaskPatrol}

The dual task domain was first introduced to
evaluate Evolvable-Substrate
HyperNEAT~\cite{risi:alife12}, 
but is appropriated here for testing
multimodal approaches.
It consists of two isolated tasks, 
hallway navigation and foraging, where 
performance and ideal behaviors in each task
are unrelated. 

In the navigation task (Figure~\ref{fig:dualTaskHallway}), 
the robot must navigate from its starting position to the 
end of a hallway using rangefinder sensors. 
In the foraging task the robot must 
visit a sequence of waypoints in order in
a rectangular room (Figure~\ref{fig:dualTaskForage}).
Four pie-slice sensors act as a compass towards 
each next waypoint.

The agent substrate in this experiment
differs from that of the patrol domains,
but matches the substrate used in the original
experiment~\cite{risi:alife12}. These robots
have ten hidden neurons and only five rangefinder
inputs. There are also four
additional inputs for the pie-slice sensors, which 
have a y-coordinate of $-1.2$ in
the substrate.


Each task has its own fitness function. 
For the navigation task, fitness is $f_{nav}$,
the normalized distance to the goal at the
end of evaluation.
For the foraging task, 
fitness is $f_{food} = \frac{n + (1 - d_f)}{4}$, 
where $n$ is the number of waypoints 
visited (maximum four) and 
$d_f$ is the normalized distance of the robot 
to the next waypoint at the end of evaluation. 
Total fitness is
the average of $f_{nav}$ and $f_{food}$. 


This fitness function is coarser than those
in the patrol domains because it does
not matter how quickly the robot reaches its goals.
The robot has 45 seconds in each task for a total
of 90 seconds per evaluation. However,
there are now only five time steps per second.

Because the isolated tasks in this domain
have no clear geometric relationship,
applying situational policy geometry becomes
somewhat arbitrary: the situation inputs are $0$ and $1$
for the hallway and foraging tasks respectively.

While the isolated nature of the tasks in this
domain enables clear exploration of the importance
of task geometry, subtasks
in real world domains are often commingled.
Thus the next domain relaxes the constraint
of task isolation.

\subsection{Two Rooms}
\label{subsection:twoRoomsPatrol}

The two rooms domain is introduced by this paper.
Like the dual task domain, it requires hallway
navigation and foraging. However, these tasks are
no longer isolated. Instead,
two large foraging rooms are separated from each
other by a convoluted 
hallway requiring navigation (Figure~\ref{fig:twoRoomDomain}).

Each room is filled with waypoints that
the robot must visit in order, while the hallway 
contains invisible breadcrumbs that
cannot be sensed, but reward 
progressing through
the hallway.
In other words, a breadcrumb is like a waypoint
in terms of fitness, but does not register
on the robot's sensors.
The robot's sensors and substrate are the
same as in the dual task domain. The pie-slice sensors
allow the robot to forage in the rooms, but the
rangefinders are crucial for navigating the hallway.

The fitness
function for this domain 
is the same as the dual task's
foraging fitness,
except that the total number of
waypoints to be visited is 15 (which includes
the breadcrumbs in the hallway).
Therefore, the fitness is
$\frac{n + (1 - d_f)}{15}$, where
$d_f$ is the normalized distance to
the next waypoint at the end of evaluation.


An aspect
of this domain that makes it more
challenging than dual task is that
evaluation ends if the
robot collides with a wall.
Normal evaluation lasts 200 seconds so 
the robot has enough time to
explore both rooms, and there are
10 time steps per second to 
give the robot extra maneuverability
in the convoluted hallway.

Because this domain integrates
hallway navigation and foraging,
multitask and situational policy geometry
use the following task division:
one brain is active when the robot is in the hallway,
and another brain is active when it is in one of
the two rooms. The situation inputs for these
tasks are $0$ for the hallway and $1$ for the rooms.

The experiments described next
test how well different approaches
to multimodal evolution perform
in the four domains described.

\section{Experimental Setup}
\label{section:ExperimentalSetup}

In each of the four domains,
30 runs each lasting 2,000 generations were conducted for all
approaches. 
Standard HyperNEAT, which has only one
module ({\tt 1M}), provided a performance baseline.
The situational policy 
geometry ({\tt SPG}) and multitask ({\tt MT})
approaches had multiple modules reflecting
the human-specified task divisions
 for each domain.
Approaches not depending on 
human-specified task divisions
include CPPNs with two ({\tt 2M}) and three ({\tt 3M}) 
preference modules, and CPPNs 
that discovered how many modules to use
through different
forms of module mutation: {\tt MM(P)},
{\tt MM(R)}, or {\tt MM(D)}.

Population sizes differed across domains
in order to conform to previous experiments.
Team patrol populations had a 
size of 500~\cite{dambrosio:iros2011},
as did lone patrol.
The population size for dual task 
was only 300~\cite{risi:alife12},
as it was also for the two rooms 
domain.

HyperNEAT parameters
were fixed across all experiments.
There was a 
20\% elitist selection rate. 
Remaining population slots
were filled equally by 
sexual offspring that did not undergo mutation
and asexual offspring that had the
following rates of mutation:
96\% chance of connection weight mutation,~3\% 
chance of connection addition, 
and 1\% chance of node addition. 
Whenever module mutation was used,
it had a 1\% chance of occurring.
The coefficients for determining 
species similarity were 1.0 for nodes and
connections and 0.1 for weights. 
The available CPPN activation
functions were the sigmoid, Gaussian, 
absolute value, and sine functions.
These parameter settings are the
same as in the original
team patrol experiment~\cite{dambrosio:iros2011}.

The experiments in MB-HyperNEAT
led to the following results.

\section{Results}
\label{section:Results}

From a high level, 
the results show that multimodal
approaches discover better
behavior faster than {\tt 1M},
and that multitask and preference neuron
approaches can evolve skilled
multimodal behavior without
any notion of situational policy
geometry. Details are
presented below.

\begin{figure*}[t!]
  \centering 
\makebox[\textwidth]{
\subfloat[Team Patrol Results]{
  \label{fig:teamPatrol}
  \includegraphics[width=0.5\textwidth]{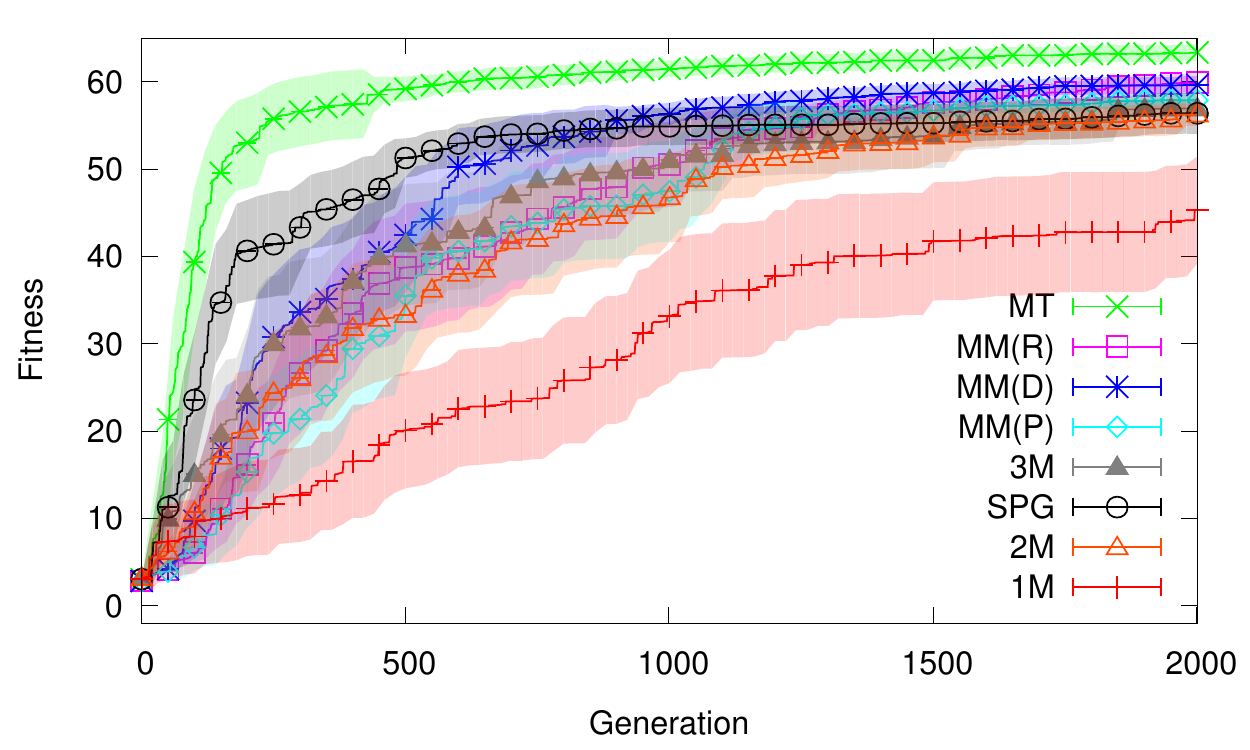}}
\subfloat[Lone Patrol Results]{
  \label{fig:lonePatrol}
  \includegraphics[width=0.5\textwidth]{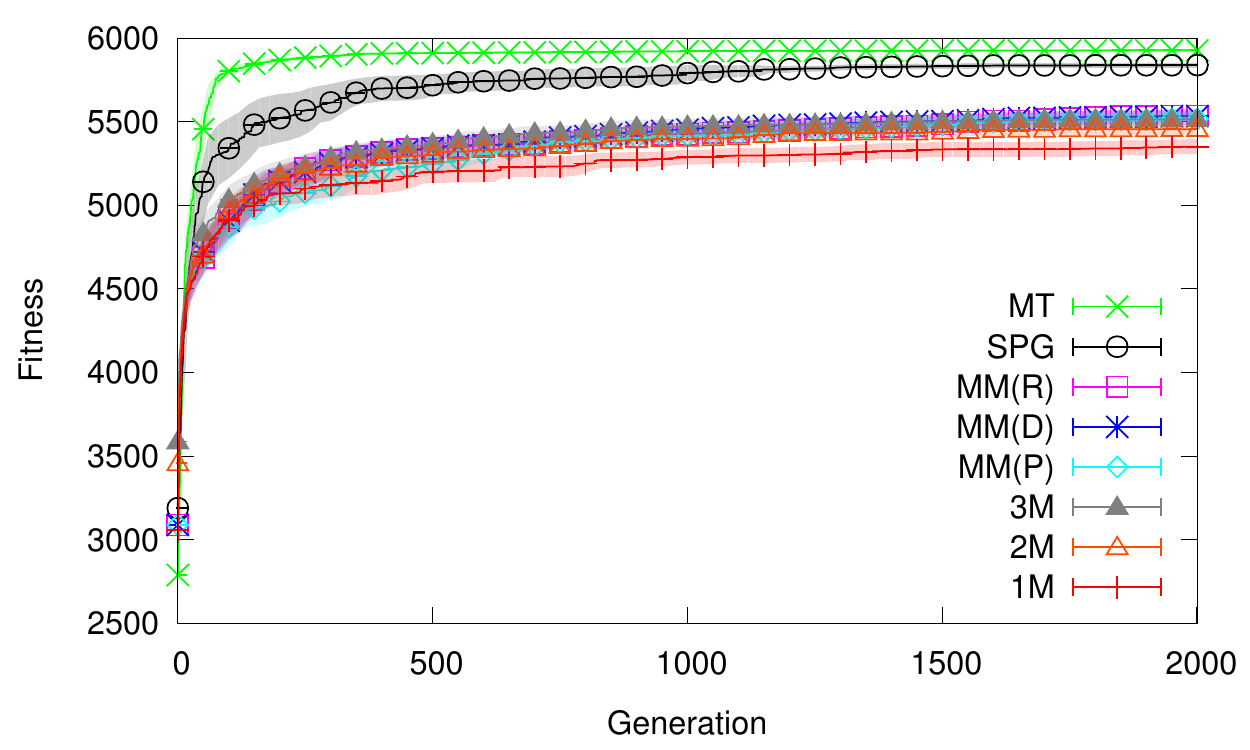}} 
}
\makebox[\textwidth]{
\subfloat[Dual Task Results]{
  \label{fig:dualTask}
  \includegraphics[width=0.5\textwidth]{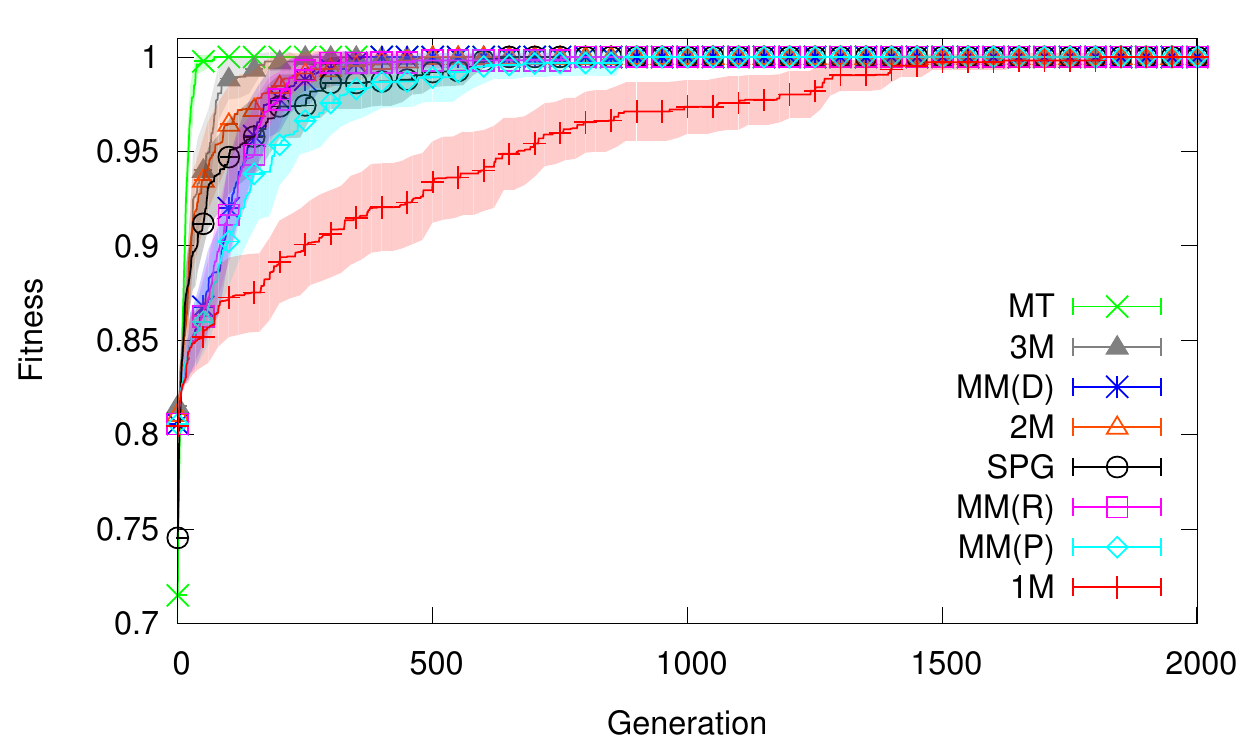}} 
\subfloat[Two Rooms Results]{
  \label{fig:twoRoom}
  \includegraphics[width=0.5\textwidth]{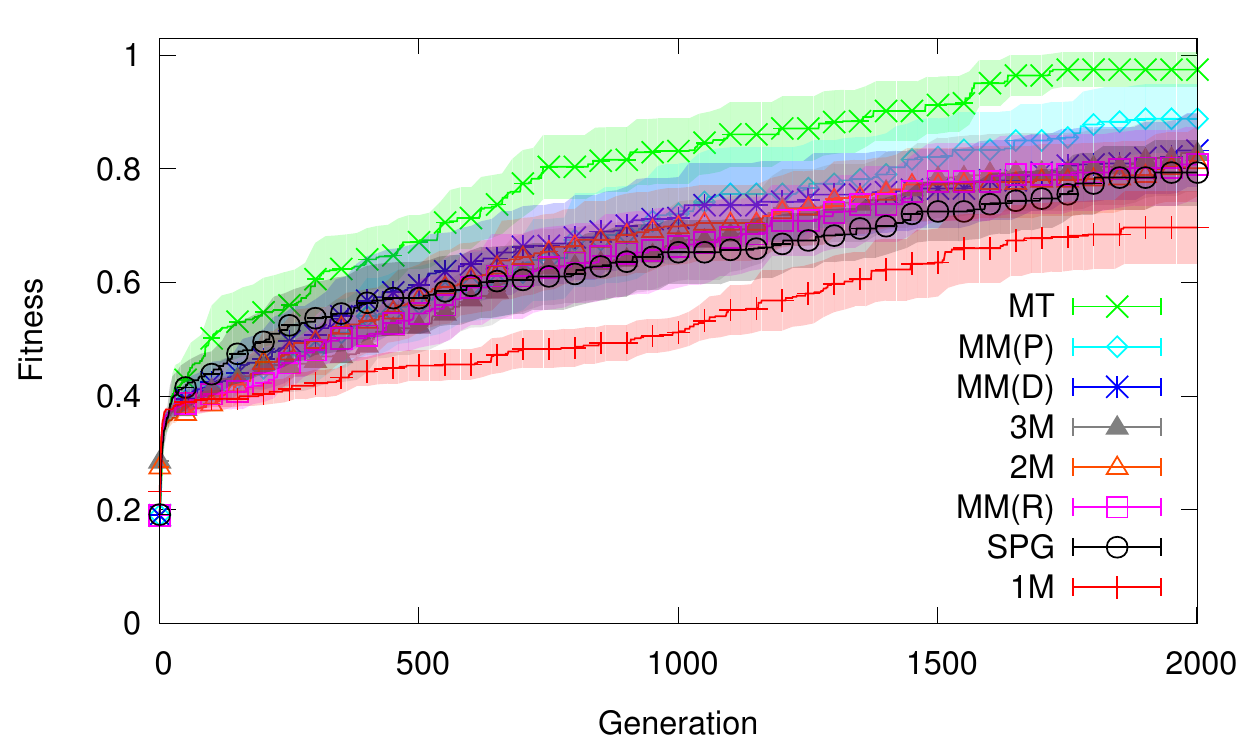}} 
} 
\caption{\small {\bf Experimental Results Across Domains.}
{\normalfont
Average champion scores across~30 runs of evolution for each approach
are shown across domains. Transparent regions show 95\%
confidence intervals. 
The key for each figure
lists methods in order of final score, with ties
broken according to which method reached the score first.
\protect\subref{fig:teamPatrol} In team patrol, 
{\tt MT} quickly outperforms all
other methods. Preference neuron approaches 
and {\tt SPG} plateau around the same 
lower score, 
but are all better than {\tt 1M}.
\protect\subref{fig:lonePatrol} In lone patrol,
confidence intervals are very narrow, indicating strong convergence
across different runs of the same method.
{\tt MT} is still the best, quickly reaching a high plateau.
{\tt SPG} is worse
than {\tt MT}, but better than all preference
neuron approaches. 
Preference neuron approaches cluster
together, but  
are slightly 
better than {\tt 1M}.
\protect\subref{fig:dualTask} The dual task domain
also has narrow confidence intervals.
All approaches plateau at the same perfect score,
but {\tt 1M} takes much longer to reach this plateau. 
{\tt MT} is still the best, because 
it reaches the plateau
earlier than other approaches. 
\protect\subref{fig:twoRoom} Modular approaches in the two rooms
domain outperform {\tt 1M}, although there is much overlap
in confidence intervals.
{\tt MM(D)}, {\tt MM(R)}, {\tt 2M}, {\tt 3M},
and {\tt SPG} cluster together and achieve nearly the same score.
{\tt MM(P)} is slightly better, and is the only method whose
final confidence intervals overlap with those of {\tt MT}, 
which achieves
a much higher score than the other methods.
In aggregate, these results show how having
multiple brains allows evolved agents to 
reach better levels of performance faster in domains 
requiring multimodal behavior.}}
\label{fig:results}
\end{figure*}

\setlength{\tabcolsep}{3pt}

\begin{table*}[htbp]
\caption{\label{tab:bonferroni}
\small Adjusted $p$-Values From Pairwise Post-hoc Mann-Whitney $U$ Tests with Bonferroni Error Correction.
{\normalfont 
In each domain, each approach is compared to every other approach
using two-tailed Mann-Whitney $U$ tests. In 
team patrol, 
lone patrol, and 
two rooms, fitness scores of champions in the final 
generation are compared. In 
dual task, all champions achieve the maximum score,
so the numbers of generations taken to reach
this score are compared.
The $p$-values of each comparison are shown.
Note that Bonferroni error
correction is used to adjust all $p$-values to prevent
spurious detection of statistical differences. 
{\bf Bolded} values indicate a significant difference
at the $p < 0.01$ level while {\it italicized} values 
(two rooms only) are for 
the $p < 0.05$ level. Values using the standard font indicate a lack
of a significant difference. Columns and rows in each table
are organized from worst to best as in the keys 
of Figure~\ref{fig:results}.
The conclusion is that {\tt MT} 
is the best in each domain while {\tt 1M} is the
worst. There are significant differences 
between other modular approaches as well.}}

\setlength\extrarowheight{1.8pt}

\small
  \begin{center}
    \begin{tabular}{|r|c|c|c|c|c|c|c|}
		\hline
\multicolumn{8}{|c|}{Team Patrol} \\
        \hline
           &{\tt 1M}
           &{\tt 2M}
           &{\tt SPG}
           &{\tt 3M}
           &{\tt MM(P)}
           &{\tt MM(D)}
           &{\tt MM(R)}\\
        \hline
{\tt 2M}&$\mathbf{0.00091}$& - & - & - &   -    & -  & -\\
        \hline
{\tt SPG}&$\mathbf{6.3\times10^{-5}}$ & 1.0 & - & - & - & - & - \\
        \hline
{\tt 3M}&$\mathbf{3.8\times10^{-7}}$ & 1.0 & 1.0 & - & - & -  & - \\
        \hline
{\tt MM(P)}&$\mathbf{4.3\times10^{-10}}$& 1.0 & 1.0 & 1.0 & - & - & - \\
        \hline
{\tt MM(D)}&$\mathbf{1.4\times10^{-11}}$&$\mathbf{0.00286}$&0.58037&$\mathbf{0.00458}$&0.17764& -  & -\\
        \hline
{\tt MM(R)}&$\mathbf{1.4\times10^{-11}}$&$\mathbf{0.00039}$&0.49246&$\mathbf{0.00458}$&0.12076&1.0& - \\
        \hline
{\tt MT}&$\mathbf{9.5\times10^{-16}}$&$\mathbf{2.1\times10^{-10}}$&$\mathbf{4.8\times10^{-7}}$&$\mathbf{2.6\times10^{-11}}$&$\mathbf{7.2\times10^{-10}}$&$\mathbf{0.00013}$&$\mathbf{0.00031}$\\
        \hline
\multicolumn{8}{|c|}{Lone Patrol} \\
		\hline

\hline
           &{\tt 1M}
           &{\tt 2M}
           &{\tt 3M}
           &{\tt MM(P)}
           &{\tt MM(D)}
           &{\tt MM(R)}
           &{\tt SPG}\\
        \hline
   {\tt 2M}&$\mathbf{0.002}$& - & - & - & - & - & -\\
        \hline
   {\tt 3M}&$\mathbf{1.6\times10^{-5}}$& 0.861 & - & - & - & - & -\\
        \hline
{\tt MM(P)}&$\mathbf{8.8\times10^{-6}}$& 0.767 & 1.0 & - & - & - & -\\
        \hline
{\tt MM(D)}&$\mathbf{2.9\times10^{-6}}$& 0.115 & 1.0 & 1.0 & - & - & -\\
        \hline
{\tt MM(R)}&$\mathbf{3.2\times10^{-6}}$& 0.162 & 1.0 & 1.0 & 1.0 & - & -\\
        \hline
  {\tt SPG}&$\mathbf{4.7\times10^{-16}}$&$\mathbf{4.7\times10^{-16}}$&$\mathbf{4.7\times10^{-16}}$&$\mathbf{4.7\times10^{-16}}$&$\mathbf{4.7\times10^{-16}}$&$\mathbf{9.5\times10^{-16}}$& -\\
        \hline
   {\tt MT}&$\mathbf{4.7\times10^{-16}}$&$\mathbf{4.7\times10^{-16}}$&$\mathbf{4.7\times10^{-16}}$&$\mathbf{4.7\times10^{-16}}$&$\mathbf{4.7\times10^{-16}}$&$\mathbf{4.7\times10^{-16}}$&$\mathbf{5.6\times10^{-12}}$\\
        \hline

\multicolumn{8}{|c|}{Dual Task} \\
		\hline


\hline
           &{\tt 1M}
           &{\tt MM(P)}
           &{\tt MM(R)}
           &{\tt SPG}
           &{\tt 2M}
           &{\tt MM(D)}
           &{\tt 3M}\\
        \hline
{\tt MM(P)}&$\mathbf{0.00104}$& - & - & - & - & - & -\\
        \hline
{\tt MM(R)}&$\mathbf{5.0\times10^{-5}}$& 0.98347 & - & - & - & - & -\\
        \hline
  {\tt SPG}&$\mathbf{7.5\times10^{-5}}$& 0.91389 & 1.0 & - & - & - & -\\
        \hline
   {\tt 2M}&$\mathbf{4.7\times10^{-6}}$&$\mathbf{0.00418}$& 0.10761 & 1.0 & - & - & -\\
        \hline
{\tt MM(D)}&$\mathbf{3.9\times10^{-5}}$& 0.60187 & 1.0 & 1.0 & 0.34905 & - & -\\
        \hline
   {\tt 3M}&$\mathbf{1.1\times10^{-6}}$&$\mathbf{9.1\times10^{-6}}$&$\mathbf{0.00012}$& 0.11535 & 1.0 &$\mathbf{0.00046}$& -\\
        \hline
   {\tt MT}&$\mathbf{3.6\times10^{-8}}$&$\mathbf{5.2\times10^{-9}}$&$\mathbf{6.6\times10^{-9}}$&$\mathbf{8.7\times10^{-6}}$&$\mathbf{0.00308}$&$\mathbf{8.4\times10^{-9}}$&$\mathbf{0.00627}$\\
        \hline

\multicolumn{8}{|c|}{Two Rooms} \\
		\hline


\hline
           &{\tt 1M}
           &{\tt SPG}
           &{\tt MM(R)}
           &{\tt 2M}
           &{\tt 3M}
           &{\tt MM(D)}
           &{\tt MM(P)}\\
        \hline
  {\tt SPG}&$\mathit{0.04858}$& - & - & - & - & - & -\\
        \hline
{\tt MM(R)}& 0.41388 & 1.0 & - & - & - & - & -\\
        \hline
   {\tt 2M}& 0.07329 & 1.0 & 1.0 & - & - & - & -\\
        \hline
   {\tt 3M}& 0.10012 & 1.0 & 1.0 & 1.0 & - & - & -\\
        \hline
{\tt MM(D)}& 0.08861 & 1.0 & 1.0 & 1.0 & 1.0 & - & -\\
        \hline
{\tt MM(P)}&$\mathbf{0.00217}$& 1.0 & 1.0 & 1.0 & 1.0 & 1.0 & -\\
        \hline
   {\tt MT}&$\mathbf{7.7\times10^{-7}}$&$\mathbf{0.0002}$&$\mathbf{0.00067}$&$\mathbf{0.00121}$&$\mathit{0.01137}$&$\mathbf{0.00366}$& 0.3591 \\
        \hline
    \end{tabular}%
  \end{center}

\end{table*}

\subsection{Team Patrol Results}
\label{subsection:teamPatrolResults}

Aligning with previous studies 
in this domain, {\tt SPG} outperforms {\tt 1M}. 
However, {\tt MT} outperforms {\tt SPG}, and preference
neuron approaches eventually reach scores
around or slightly above those of 
{\tt SPG} (Figure~\ref{fig:teamPatrol}).

In the final generation, the Kruskal-Wallis
test indicates a significant difference
between champion fitness scores of different approaches 
($H=117.7784,df=7,N=30,p < 2.2 \times 10^{-16}$).
Post-hoc tests indicate that
all modular approaches significantly
outperform {\tt 1M}, and {\tt MT}
significantly outperforms all other methods.
These differences and differences between
some preference neuron methods are
reported in Table~\ref{tab:bonferroni}. 



Observation of evolved behaviors
reveals qualitative differences between
methods. The behavior of {\tt MT}
networks involves each robot going directly
to its destination and returning
in perfect synchronicity. Less skilled modular networks
will generally have small inefficiencies,
such as one out of the three robots lagging
slightly behind the others. In the worst
runs, one of the robots becomes
stuck advancing outward and
fails to return. Such failure is common in
{\tt 1M} runs, but happens to some preference
neuron champions as well. Videos of several
representative behaviors are at
\url{southwestern.edu/~schrum2/re/team-patrol.html}.

These videos also reveal how preference
neuron approaches switch
between brains. 
Interestingly, 
different team members switch brains
at different times.
Some robots use 
a single brain for
advancing and returning, while others
in the same team frequently switch 
brains. Most common is to
rely on one brain to advance, 
switch to turn around once the signal input 
activates, and then switch back to the original
brain to return home.

Module mutation champions 
produce many unused brains. 
Some have 
between 10 and 20 substrate brains, 
but agents use no more 
than three of them. This
result occurs in the other domains
as well.

\subsection{Lone Patrol Results}
\label{subsection:lonePatrolResults}

Results in the lone patrol domain 
have little variation; 
runs of each method
quickly converge (Figure~\ref{fig:lonePatrol}).
{\tt MT} performs best, followed by {\tt SPG},
then all preference neuron methods 
(which cluster together), and finally 
{\tt 1M}, which performs worst.

The Kruskal-Wallis test again indicates 
significant differences between final champion
scores in this domain 
($H=162.3146,df=7,N=30,p < 2.2 \times 10^{-16}$).
In fact, post-hoc tests indicate that all
methods are significantly different from each
other, \emph{except}
methods that use preference neurons 
(Table~\ref{tab:bonferroni}). 
Specifically, {\tt 2M}, {\tt 3M}, {\tt MM(P)},
{\tt MM(R)}, and {\tt MM(D)} are not significantly
different from each other, but are different
from the other methods.


Videos of evolved behavior
show that most champions  
reach or nearly reach all waypoints.
Fitness score differences depend on route efficiency. 
Preference neuron approaches and {\tt 1M} 
perform worse because some  
champions become stuck on a corner 
when returning home after visiting 
the final waypoint.
Representative 
videos can be viewed at
\url{southwestern.edu/~schrum2/re/lone-patrol.html}.

{\tt MT} performs well because 
it proceeds directly to each waypoint
and promptly turns to the
next waypoint once the preceding one is reached.
{\tt SPG} generally does the same, but
sometimes goes around corners less efficiently than {\tt MT}.
Preference neuron networks often
waste time at each end of the plus
sign and when turning corners.
They often proceed to the end of
each hallway rather than turning around 
directly as each waypoint is reached
(they do not sense when they reach each waypoint),
and often move around corners
in discontinuous starts and stops rather than in
a continuous arc. Dealing with turns generally
requires dedicated brains. Specifically,
turning around at each dead end is often
assigned a dedicated brain, and there is often
a brain dedicated to handling turning
in the plus's center. Sometimes
turning is accomplished by thrashing
between the dedicated
turning brain and whatever brain
was being used previously. These behaviors
typically require the use of three brains,
though module mutation frequently produces
CPPNs with 10 or more modules, whose brains
are mostly unused.

Deciding on actions
when at the center of the plus sign provides
a challenge for the robot,
because it looks the same
to the robot's sensors on each visit 
(the domain is partially
observable~\cite{kaelbling:ai1998}), 
yet each visit a different
behavior is needed. 
The {\tt 1M}
networks perform the worst
because even when they successfully
visit all waypoints and return home,
they tend to navigate the plus sign's center
with an inefficient trick. 
Instead of turning
right, the robot
loops around by turning left
repeatedly. The added time
required to perform this maneuver has a large
fitness cost.

Both patrol domains have fitness functions
that not only reward reaching certain goals,
but doing so quickly. Distinctions between
methods in lone patrol depend more on how
quickly all waypoints are visited 
than on whether they are reached. However,
the fitness functions for the next two
domains only measure whether or not the
robot ever achieves its goals, irrespective
of speed.

\subsection{Dual Task Results}
\label{subsection:dualTaskPatrolResults}

All methods eventually master the
dual task, obtaining a perfect fitness 
(Figure~\ref{fig:dualTask}). However,
there are distinctions 
in how many generations each method
requires to succeed.
{\tt MT} is still the best, succeeding
in less than 100 generations 
in all runs. {\tt SPG} and the 
preference neuron methods
take slightly longer to 
reach  maximum fitness, while
{\tt 1M} takes significantly longer.





Because all runs achieve
maximum fitness, the Kruskal-Wallis
test is applied to the number of generations
necessary to succeed in this way,
and indicates
a significant difference 
($H=114.2459,df=7,N=30,p < 2.2 \times 10^{-16}$).
Post-hoc tests indicate that
all other methods succeed
significantly faster 
than {\tt 1M}, and {\tt MT}
succeeds significantly faster than
all other methods. 
These differences and others are
reported in Table~\ref{tab:bonferroni}. 


Although all methods eventually
achieve perfect fitness, there
are many examples of inefficient behavior,
because there is no selection for efficiency. 
Thus, many champions collide with 
walls in the hallway task or become stuck
for long periods. There is less time
for mistakes in the foraging task,
so performance in this task is relatively
straightforward: robots go directly to each
waypoint in sequence. Example behaviors
can be seen at
\url{southwestern.edu/~schrum2/re/dual-task.html}.

Preference neuron approaches 
actively use at most three brains. 
Instead of dedicating separate brains 
to individual tasks, as in {\tt MT} and {\tt SPG}, 
the same set of brains is repurposed across tasks.
In the hallway task, it is common for one
brain to handle moving straight forward,
while one or two others handle turning around
corners. In the foraging task it is common to
have one brain that moves straight toward the
next waypoint, while another brain takes over
to turn the robot around after each waypoint
is reached.

Although these usage patterns are common,
there are also module mutation champions
that have nearly 20 brains, yet 
solve both tasks using only one.
Many module mutation champions
use two or three brains instead, but
also leave many brains unused.


The next domain, which is closely related
to this one, nevertheless produces
very different results, as shown next.

\subsection{Two Rooms Results}
\label{subsection:twoRoomsPatrolResults}

As in the other domains, {\tt MT} performs best in two rooms, 
although its margin of success is not as
dramatic (Figure~\ref{fig:twoRoom}).
The {\tt 1M} approach is once again the worst,
with {\tt SPG} and preference neuron approaches clustering
between {\tt 1M} and {\tt MT}.

Because no method succeeds in
all runs, the Kruskal-Wallis test is again
applied to fitness scores
of the final champions, and indicates
that there are significant differences between methods
($H=43.9804,df=7,N=30,p \approx 2.156 \times 10^{-7}$).
Post-hoc tests indicate that
although {\tt 1M} has the lowest final scores,
it is only \emph{significantly} outperformed by 
{\tt SPG}, {\tt MM(P)} and {\tt MT}. 
{\tt MT} significantly outperforms all methods, 
except {\tt MM(P)} (Table~\ref{tab:bonferroni}). 


The worst-performing champions visit 
all waypoints in the first room, but
fail to progress through the hallway.
There are also slightly
more successful robots that use
foraging behavior so
inefficient that they cannot  
visit all waypoints in the second
room within the time limit, despite
successfully navigating the hallway.
Examples of inefficient behavior include
heading toward each waypoint in
tight spirals instead of in a straight line,
and circling the wall of the first room
to find the hallway instead of
directly heading to it.
This range of behaviors explains the 
broad dispersion of 
champion fitness scores. Videos of 
representative behaviors can be seen at
\url{southwestern.edu/~schrum2/re/two-rooms.html}.

When preference neurons are used,
often one brain is
mostly responsible for navigating
the hallway. However, this
brain will typically alternate rapidly
with whatever brain was previously active.
Visiting waypoints in each room is
generally handled by a single
brain, though sometimes another
brain activates to reorient the robot
after each waypoint is reached 
(as in the foraging environment of the dual task).
These types of behaviors seldom
require more than two or three brains,
and once again many
module mutation runs produce many
unnecessary brains.

Interestingly, while {\tt MT} performs best,
the decision to dedicate
a brain to hallway navigation requires
human insight. However, {\tt SPG}
performs poorly despite using the same
task division, thus providing a clear
example of how constraining different
controllers to be geometrically related
can be harmful.

\section{Discussion and Future Work}
\label{section:Discussion}

All multi-brain approaches to
creating agents are
superior to {\tt 1M} in at least
three of the four explored domains.
{\tt SPG} is superior to {\tt 1M} in all domains,
but also inferior to {\tt MT}
in all domains, thus demonstrating that
even in domains where situational policy geometry 
seems appropriate,
it is better to allow a multitask
CPPN to create completely distinct brains
instead. For this reason, if a human-specified
task division is available, multitask
CPPNs seem the most principled first approach.

However, preference neuron approaches
can be applied even when a task
division is not available, and
at least one preference neuron approach
is significantly better than {\tt 1M} in each domain.
However, because effective task divisions
were available, preference neurons are never
significantly better than {\tt SPG},
and are inferior to {\tt MT}. 
This result contrasts with previous results in 
Ms.\ Pac-Man using the direct encoding 
MM-NEAT~\cite{schrum:tciaig16,schrum:gecco2015}.
It is possible that
preference neurons are less
effective when combined with HyperNEAT 
than with directly encoded neural networks, but
it is more likely that the increased
complexity of Ms.\ Pac-Man (compared to the domains
of this paper) is what allowed preference neurons
to shine.
Therefore, applying MB-HyperNEAT to more complex
domains lacking a clear task division 
is one avenue of future work.

There is no consistent relationship
between the performance of a fixed number of
preference neuron brains 
({\tt 2M} and {\tt 3M}) and variants of module
mutation. In team patrol, module
mutation outperforms a fixed number
of brains, but this relation is
reversed in dual task. In lone patrol,
all preference neuron approaches
perform equivalently, as is the case in
two rooms (even though {\tt MM(P)}
significantly outperforms {\tt 1M}). 
Distinctions between
different forms of module mutation form
no consistent pattern either.

Even when there are statistically significant differences
between preference neuron
approaches, the effect size is small.
Observing how many brains
are actually used in each domain provides an explanation: 
regardless of how many
brains module mutation produces, the
final champions only use one to three
of them. Therefore, it makes sense that
the behaviors exhibited and fitness scores achieved
by module mutation are similar to those of
a set number of preference neuron brains.
Given the performance of {\tt MT} in all
domains, it is clear that two to three
brains are sufficient, if they are used
correctly. A domain complex enough to either
require more than three modules, or one
lacking an obvious human-specified task
division, is
likely required for 
module mutation to show practical
benefit over simpler methods.

It is unclear how 
the many unused controllers
produced by module mutation affect evolution.
Why do certain forms of
module mutation sometimes do better
than approaches with a set number of
preference neuron brains despite
producing champions that use the same
number of brains? Intuitively,
wasting mutation operations on
CPPN modules which only affect brains
that are never used seems like it
would slow down evolution. 
Selection cannot act on unused modules.
These portions of the CPPN are
effectively introns, a biological
phenomenon that is also known in the
Genetic Programming community~\cite{nordin:gp96}.

An intron is a gene that does not affect
the phenotype.
In general, introns can be safely modified without
changing a genotype's fitness.
Therefore, genotypes that already have high fitness 
are more likely to persist in the population,
because introns make them less vulnerable to
destructive mutations by providing a portion of the
genotype that can be changed without effect.
With regard to module mutation specifically,
there is also a chance that mutations in an intron
could cause long unused modules
to suddenly start being used. Such
modules will likely hurt fitness in most 
cases because they have not been subject to
any selection pressure. 
However, the sudden emergence of a good module
could help a population escape a local optimum
in the fitness landscape.
The few cases where such positive mutations occur
could be enough to make
module mutation beneficial overall, at 
least in certain domains. The performance
of {\tt MM(P)} in two rooms is an example.

This paper generated multimodal
behavior by creating several complete
brains in separate substrates.
However, HyperNEAT can also take advantage
of multiple \emph{sensory} modalities using a
multi-spatial substrate 
(MSS~\cite{pugh:gecco2013}),
which can embed the neurons of
a single brain into several
sub-substrates. Different
modalities of input are separated into 
different substrates that are integrated by
a hidden layer substrate, which eventually
propagates to a final output substrate. The MSS approach
could be compared against, and even
combined with the methods of this paper
in future work.

Another focus for future work
is the evolution of larger 
networks. One of the primary
benefits of HyperNEAT is its
ability to compactly encode large
networks~\cite{hausknecht:tciaig14,gauci:aaai08,gauci:ppsn2010}, 
so it is important to verify that
the HyperNEAT extensions presented
in this paper also provide
a benefit to the large, complex
networks that HyperNEAT was designed to create.

\section{Conclusion}
\label{section:Conclusion}

Automatic and effective 
evolution of complex multimodal
behavior requires indirect
encodings and mechanisms
that support 
evolving distinct neural
structures. The main
idea in this paper is to
combine the popular HyperNEAT indirect
encoding and the MM-NEAT
approach to evolving
modular networks, thereby 
realizing the strengths of both
approaches.
The result is MB-HyperNEAT,
a collection of methods for creating
multiple brains for a single agent.
Results show that the multitask
CPPN approach
always outperforms a previous
attempt to merge HyperNEAT
with multimodal extensions known as
situational policy geometry,
and that approaches using
preference neurons make
it possible to
evolve agents with multiple brains when
a human-specified task 
division is unavailable.
Preference neuron approaches
achieve lower scores than multitask CPPNs,
but even though they
are not provided with a human-specified
task division, their scores are often
statistically tied with those of situational
policy geometry,
and generally surpass scores of 
agents with only
one brain. The conclusion is that
MB-HyperNEAT is a promising
toolkit for evolving
complex multimodal behavior
that can reduce the need
for specialized domain knowledge.





\ifCLASSOPTIONcaptionsoff
  \newpage
\fi



%

\bibliographystyle{IEEEtran}
\bibliography{MB-HyperNEAT}

\begin{thebibliography}{10}
\providecommand{\url}[1]{#1}
\csname url@samestyle\endcsname
\providecommand{\newblock}{\relax}
\providecommand{\bibinfo}[2]{#2}
\providecommand{\BIBentrySTDinterwordspacing}{\spaceskip=0pt\relax}
\providecommand{\BIBentryALTinterwordstretchfactor}{4}
\providecommand{\BIBentryALTinterwordspacing}{\spaceskip=\fontdimen2\font plus
\BIBentryALTinterwordstretchfactor\fontdimen3\font minus
  \fontdimen4\font\relax}
\providecommand{\BIBforeignlanguage}[2]{{%
\expandafter\ifx\csname l@#1\endcsname\relax
\typeout{** WARNING: IEEEtran.bst: No hyphenation pattern has been}%
\typeout{** loaded for the language `#1'. Using the pattern for}%
\typeout{** the default language instead.}%
\else
\language=\csname l@#1\endcsname
\fi
#2}}
\providecommand{\BIBdecl}{\relax}
\BIBdecl

\bibitem{local:stanley:ec02}
K.~O. Stanley and R.~Miikkulainen, ``{E}volving {N}eural {N}etworks {T}hrough
  {A}ugmenting {T}opologies,'' \emph{Evolutionary Computation Journal},
  vol.~10, pp. 99--127, 2002.

\bibitem{floreano2008neuroevolution}
D.~Floreano, P.~D{\"u}rr, and C.~Mattiussi, ``Neuroevolution: {F}rom
  {A}rchitectures to {L}earning,'' \emph{Evolutionary Intelligence}, vol.~1,
  no.~1, pp. 47--62, 2008.

\bibitem{local:calabretta:alife00}
R.~Calabretta, S.~Nolfi, D.~Parisi, and G.~Wagner, ``{D}uplication of {M}odules
  {F}acilitates the {E}volution of {F}unctional {S}pecialization,''
  \emph{Artificial Life}, vol.~6, no.~1, pp. 69--84, 2000.

\bibitem{schrum:tciaig16}
\BIBentryALTinterwordspacing
J.~Schrum and R.~Miikkulainen, ``{Discovering Multimodal Behavior in Ms.
  Pac-Man through Evolution of Modular Neural Networks},'' \emph{TCIAIG},
  vol.~8, no.~1, pp. 67--81, 2016. [Online]. Available:
  \url{http://nn.cs.utexas.edu/?schrum:tciaig16}
\BIBentrySTDinterwordspacing

\bibitem{stanley:alife2009}
\BIBentryALTinterwordspacing
K.~O. Stanley, D.~B. D'Ambrosio, and J.~Gauci, ``A {H}ypercube-based {E}ncoding
  for {E}volving {L}arge-scale {N}eural {N}etworks,'' \emph{Artificial Life},
  vol.~15, no.~2, pp. 185--212, Apr. 2009. [Online]. Available:
  \url{http://dx.doi.org/10.1162/artl.2009.15.2.15202}
\BIBentrySTDinterwordspacing

\bibitem{dambrosio:iros2011}
D.~B. D'Ambrosio, J.~Lehman, S.~Risi, and K.~O. Stanley, ``{Task Switching in
  Multirobot Learning Through Indirect Encoding},'' in \emph{International
  Conference on Intelligent Robots and Systems}.\hskip 1em plus 0.5em minus
  0.4em\relax IEEE, 2011, pp. 2802--2809.

\bibitem{schrum:gecco14}
\BIBentryALTinterwordspacing
J.~Schrum and R.~Miikkulainen, ``{E}volving {M}ultimodal {B}ehavior {W}ith
  {M}odular {N}eural {N}etworks in {M}s.\ {P}ac-{M}an,'' in \emph{Genetic and
  Evolutionary Computation Conference}.\hskip 1em plus 0.5em minus 0.4em\relax
  ACM, July 2014, pp. 325--332. [Online]. Available:
  \url{http://nn.cs.utexas.edu/?schrum:gecco2014}
\BIBentrySTDinterwordspacing

\bibitem{schrum:tciaig12}
\BIBentryALTinterwordspacing
------, ``{Evolving Multimodal Networks for Multitask Games},'' \emph{IEEE
  Transactions on Computational Intelligence and AI in Games}, vol.~4, no.~2,
  pp. 94--111, June 2012. [Online]. Available:
  \url{http://nn.cs.utexas.edu/?schrum:tciaig12}
\BIBentrySTDinterwordspacing

\bibitem{hausknecht:tciaig14}
\BIBentryALTinterwordspacing
M.~Hausknecht, J.~Lehman, R.~Miikkulainen, and P.~Stone, ``{A Neuroevolution
  Approach to General Atari Game Playing},'' \emph{IEEE Transactions on
  Computational Intelligence and AI in Games}, vol.~6, no.~4, pp. 355--366, Dec
  2014. [Online]. Available: \url{http://nn.cs.utexas.edu/?hausknecht:tciaig14}
\BIBentrySTDinterwordspacing

\bibitem{gauci:aaai08}
J.~Gauci and K.~O. Stanley, ``{A Case Study on the Critical Role of Geometric
  Regularity in Machine Learning},'' in \emph{National Conference on Artificial
  Intelligence (AAAI-08)}.\hskip 1em plus 0.5em minus 0.4em\relax AAAI Press,
  2008, pp. 628--633.

\bibitem{gauci:ppsn2010}
\BIBentryALTinterwordspacing
------, \emph{Parallel Problem Solving from Nature}.\hskip 1em plus 0.5em minus
  0.4em\relax Berlin, Heidelberg: Springer Berlin Heidelberg, 2010, ch.
  {Indirect Encoding of Neural Networks for Scalable Go}, pp. 354--363.
  [Online]. Available: \url{http://dx.doi.org/10.1007/978-3-642-15844-5_36}
\BIBentrySTDinterwordspacing

\bibitem{haykin:book99}
S.~Haykin, \emph{Neural Networks, A Comprehensive Foundation}.\hskip 1em plus
  0.5em minus 0.4em\relax Upper Saddle River, New Jersey: Prentice Hall, 1999.

\bibitem{kashtan:nasusa2005}
N.~Kashtan and U.~Alon, ``{Spontaneous Evolution of Modularity and Network
  Motifs},'' \emph{National Academy of Sciences USA}, vol. 102, no.~39, pp.
  13\,773--13\,778, 2005.

\bibitem{clune:royal2013}
J.~Clune, J.-B. Mouret, and H.~Lipson, ``{The Evolutionary Origins of
  Modularity},'' \emph{Royal Society B: Biological Sciences}, vol. 280, no.
  1755, pp. 20\,122\,863--20\,122\,863, 2013.

\bibitem{huizinga:gecco14}
J.~Huizinga, J.-B. Mouret, and J.~Clune, ``{Evolving Neural Networks That Are
  Both Modular and Regular: HyperNEAT Plus the Connection Cost Technique},'' in
  \emph{Genetic and Evolutionary Computation Conference}.\hskip 1em plus 0.5em
  minus 0.4em\relax ACM, July 2014, pp. 697--704.

\bibitem{nolfi:adaptivebehavior96}
S.~Nolfi, ``{Using Emergent Modularity to Develop Control Systems for Mobile
  Robots},'' \emph{Adaptive Behavior}, vol.~5, pp. 343--363, 1996.

\bibitem{schrum:gecco2015}
\BIBentryALTinterwordspacing
J.~Schrum and R.~Miikkulainen, ``{S}olving {I}nterleaved and {B}lended
  {S}equential {D}ecision-{M}aking {P}roblems through {M}odular
  {N}euroevolution,'' in \emph{Genetic and Evolutionary Computation
  Conference}.\hskip 1em plus 0.5em minus 0.4em\relax ACM, July 2015, pp.
  345--352. [Online]. Available:
  \url{http://nn.cs.utexas.edu/?schrum:gecco2015}
\BIBentrySTDinterwordspacing

\bibitem{howard:cec2010}
G.~Howard, L.~Bull, and P.-L. Lanzi, ``{A Spiking Neural Representation for
  XCSF},'' in \emph{Congress on Evolutionary Computation}.\hskip 1em plus 0.5em
  minus 0.4em\relax IEEE, 2010, pp. 1--8.

\bibitem{hurst:alife06}
\BIBentryALTinterwordspacing
J.~Hurst and L.~Bull, ``{A Neural Learning Classifier System with Self-Adaptive
  Constructivism for Mobile Robot Control},'' \emph{Artificial Life}, vol.~12,
  no.~3, pp. 353--380, 2006. [Online]. Available:
  \url{http://dx.doi.org/10.1162/artl.2006.12.3.353}
\BIBentrySTDinterwordspacing

\bibitem{dam:tkde2008}
\BIBentryALTinterwordspacing
H.~H. Dam, H.~A. Abbass, and C.~Lokan, ``Neural-based learning classifier
  systems,'' \emph{IEEE Transactions on Knowledge and Data Engineering},
  vol.~20, no.~1, pp. 26--39, 2008. [Online]. Available:
  \url{http://ieeexplore.ieee.org/lpdocs/epic03/wrapper.htm?arnumber=4358957}
\BIBentrySTDinterwordspacing

\bibitem{togelius:jifs04}
J.~Togelius, ``{Evolution of a Subsumption Architecture Neurocontroller},''
  \emph{Intelligent and Fuzzy Systems}, pp. 15--20, 2004.

\bibitem{thompson:cig2009}
T.~Thompson, F.~Milne, A.~Andrew, and J.~Levine, ``Improving {C}ontrol
  {T}hrough {S}ubsumption in the {E}vo{T}anks {D}omain,'' in \emph{Conference
  on Computational Intelligence and Games}.\hskip 1em plus 0.5em minus
  0.4em\relax IEEE, 2009, pp. 363--370.

\bibitem{vanhoorn:cig2009}
N.~van Hoorn, J.~Togelius, and J.~Schmidhuber, ``{H}ierarchical {C}ontroller
  {L}earning in a {F}irst-{P}erson {S}hooter,'' in \emph{Conference on
  Computational Intelligence and Games}.\hskip 1em plus 0.5em minus 0.4em\relax
  IEEE, 2009, pp. 294--301.

\bibitem{lessin:gecco13}
\BIBentryALTinterwordspacing
D.~Lessin, D.~Fussell, and R.~Miikkulainen, ``{Open-Ended Behavioral Complexity
  for Evolved Virtual Creatures},'' in \emph{Genetic and Evolutionary
  Computation Conference}.\hskip 1em plus 0.5em minus 0.4em\relax ACM, 2013,
  pp. 335--342. [Online]. Available:
  \url{http://nn.cs.utexas.edu/?lessin:gecco13}
\BIBentrySTDinterwordspacing

\bibitem{lessin:alife14}
\BIBentryALTinterwordspacing
------, ``{Adapting Morphology to Multiple Tasks in Evolved Virtual
  Creatures},'' in \emph{International Conference on the Synthesis and
  Simulation of Living Systems (ALIFE 14)}, 2014. [Online]. Available:
  \url{http://nn.cs.utexas.edu/?lessin:alife14}
\BIBentrySTDinterwordspacing

\bibitem{caruana:icml93}
R.~A. Caruana, ``{Multitask Learning: A Knowledge-based Source of Inductive
  Bias},'' in \emph{International Conference on Machine Learning}, 1993, pp.
  41--48.

\bibitem{secretan:ecj2011}
\BIBentryALTinterwordspacing
J.~Secretan, N.~Beato, D.~B. D'Ambrosio, A.~Rodriguez, A.~Campbell, J.~T.
  Folsom-Kovarik, and K.~O. Stanley, ``{Picbreeder: A Case Study in
  Collaborative Evolutionary Exploration of Design Space},'' \emph{Evolutionary
  Computation}, vol.~19, no.~3, pp. 373--403, Sep. 2011. [Online]. Available:
  \url{http://dx.doi.org/10.1162/EVCO_a_00030}
\BIBentrySTDinterwordspacing

\bibitem{risi:alife12}
S.~Risi and K.~O. Stanley, ``{An Enhanced Hypercube-Based Encoding for Evolving
  the Placement, Density and Connectivity of Neurons},'' \emph{Artificial
  Life}, vol.~18, no.~4, pp. 331--363, 2012.

\bibitem{drchal:icann2009}
J.~Drchal, O.~Kapr{\'a}l, J.~Koutn{\'i}k, and M.~{\v S}norek,
  ``\BIBforeignlanguage{English}{Combining {M}ultiple {I}nputs in {HyperNEAT}
  {M}obile {A}gent {C}ontroller},'' in
  \emph{\BIBforeignlanguage{English}{International Conference on Artificial
  Neural Networks, Part II}}, vol.~2.\hskip 1em plus 0.5em minus 0.4em\relax
  Berlin: Springer, 2009, pp. 775--783.

\bibitem{pugh:gecco2013}
\BIBentryALTinterwordspacing
J.~K. Pugh and K.~O. Stanley, ``{Evolving Multimodal Controllers with
  HyperNEAT},'' in \emph{Genetic and Evolutionary Computation
  Conference}.\hskip 1em plus 0.5em minus 0.4em\relax ACM, 2013, pp. 735--742.
  [Online]. Available: \url{http://doi.acm.org/10.1145/2463372.2463459}
\BIBentrySTDinterwordspacing

\bibitem{dambrosio:aamas2010}
D.~B. D'Ambrosio, J.~Lehman, S.~Risi, and K.~O. Stanley, ``{Evolving Policy
  Geometry for Scalable Multiagent Learning},'' in \emph{Proceedings of the
  International Conference on Autonomous Agents and Multiagent Systems
  (AAMAS)}, 2010, pp. 731--738.

\bibitem{kaelbling:ai1998}
L.~P. Kaelbling, M.~L. Littman, and A.~R. Cassandra, ``{Planning and Acting in
  Partially Observable Stochastic Domains},'' \emph{Artificial intelligence},
  vol. 101, no.~1, pp. 99--134, 1998.

\bibitem{nordin:gp96}
\BIBentryALTinterwordspacing
P.~Nordin, F.~Francone, and W.~Banzhaf, ``{Explicitly Defined Introns and
  Destructive Crossover in Genetic Programming},'' in \emph{{Advances in
  Genetic Programming}}.\hskip 1em plus 0.5em minus 0.4em\relax MIT Press,
  1996, pp. 111--134. [Online]. Available:
  \url{http://dl.acm.org/citation.cfm?id=270195.270205}
\BIBentrySTDinterwordspacing

\end{thebibliography}

%
%

%








\end{document}